%% file: main.tex
\documentclass{article}

\usepackage[preprint]{neurips_2026}

\usepackage[utf8]{inputenc}
\usepackage[T1]{fontenc}
\usepackage{hyperref}
\usepackage{url}
\usepackage{booktabs}
\usepackage{amsfonts}
\usepackage{amsmath}
\usepackage{amssymb}
\usepackage{nicefrac}
\usepackage{microtype}
\usepackage{xcolor}
\usepackage{graphicx}
\usepackage{subcaption}
\usepackage{enumitem}
\usepackage{multirow}
\usepackage{xspace}
\usepackage{pifont}
\usepackage{wrapfig}
\usepackage{listings}
\usepackage{tcolorbox}
\tcbuselibrary{listings,skins,breakable}

\definecolor{goldenBg}{HTML}{E8F5E9}
\definecolor{hallucBg}{HTML}{FFEBEE}
\definecolor{promptBg}{HTML}{F3E5F5}
\definecolor{codeBg}{HTML}{F5F5F5}
\definecolor{dockerBg}{HTML}{E3F2FD}
\tcbset{
  goldenstyle/.style={
    colback=goldenBg, colframe=green!60!black,
    fonttitle=\bfseries\small,
    title={\ding{51}~Golden Completion},
    listing only,
    left=4pt, right=4pt, top=2pt, bottom=2pt,
    boxrule=0.5pt, arc=2pt,
    listing options={
      basicstyle=\ttfamily\footnotesize,
      breaklines=true,
      columns=fullflexible
    }
  },
  hallucstyle/.style={
    colback=hallucBg, colframe=red!60!black,
    fonttitle=\bfseries\small,
    title={\ding{55}~Hallucinated Completion},
    listing only,
    left=4pt, right=4pt, top=2pt, bottom=2pt,
    boxrule=0.5pt, arc=2pt,
    listing options={
      basicstyle=\ttfamily\footnotesize,
      breaklines=true,
      columns=fullflexible
    }
  },
  promptstyle/.style={
    colback=promptBg, colframe=purple!50!black,
    fonttitle=\bfseries\small,
    listing only,
    left=4pt, right=4pt, top=2pt, bottom=2pt,
    boxrule=0.5pt, arc=2pt,
    listing options={
      basicstyle=\ttfamily\footnotesize,
      breaklines=true,
      columns=fullflexible
    }
  },
  codestyle/.style={
    colback=codeBg, colframe=gray!50!black,
    fonttitle=\bfseries\small,
    listing only,
    left=4pt, right=4pt, top=2pt, bottom=2pt,
    boxrule=0.5pt, arc=2pt,
    listing options={
      basicstyle=\ttfamily\footnotesize,
      breaklines=true,
      columns=fullflexible
    }
  },
dockerstyle/.style={
  listing engine=listings,   
  colback=dockerBg,
  colframe=blue!50!black,
  fonttitle=\bfseries\small,
  title={Docker Commands},
  listing only,
  left=4pt, right=4pt,
  top=2pt, bottom=2pt,
  boxrule=0.5pt, arc=2pt,
  listing options={
    language=bash,
    basicstyle=\ttfamily\footnotesize,
    breaklines=true
  }
}
}

\newtcblisting{goldenbox}[1][]{goldenstyle,#1}
\newtcblisting{hallucbox}[1][]{hallucstyle,#1}

\newtcblisting{promptbox}[2][]{promptstyle,title={#2},#1}
\newtcblisting{codebox}[2][]{codestyle,title={#2},#1}

\newtcblisting{dockerbox}[1][]{dockerstyle,#1}

\newcommand{\benchmark}{\textsc{Delulu}\xspace}
\newcommand{\claude}{\textsc{Claude Sonnet}\xspace}
\newcommand{\gptfiveone}{\textsc{GPT-5.1}\xspace}
\newcommand{\gptfivetwo}{\textsc{GPT-5.2}\xspace}
\newcommand{\gptfivetwocodex}{\textsc{GPT-5.2-Codex}\xspace}
\newcommand{\glm}{\textsc{GLM-4.7}\xspace}
\newcommand{\cmark}{\ding{51}}
\newcommand{\xmark}{\ding{55}}

\newif\ifpreprint
\preprinttrue 
\newlength{\savespaceunit}
\newcommand{\savespace}[1][1]{%
  \ifpreprint\else\vspace{#1\savespaceunit}\fi
}

\title{\benchmark: A Verified Multi-Lingual Benchmark for Code Hallucination Detection in Fill-in-the-Middle Tasks}

\author{
  Mahdi Erfanian\thanks{Work done during a project at Microsoft CoreAI.} \\University of Illinois Chicago \And
  Nelson Daniel Troncoso \\Microsoft \And
  Aashna Garg \\Microsoft
  \AND
  Amabel Gale\\Microsoft \And
  Xiaoyu Liu \\Microsoft \And
  Pareesa Ameneh Golnari \\Microsoft \And
  Shengyu Fu \\Microsoft \AND
  \texttt{merfan2@uic.edu} \quad \texttt{\{ntroncoso, aashnagarg, amabelgale} \\ \texttt{xiaoyu.liu, pareesa.golnari, shengyfu\}@microsoft.com} \\
}

\begin{document}
\maketitle
\savespace[1.2]
\begin{abstract}
\savespace[0.5]
Large Language Models for code generation frequently produce \emph{hallucinations} in Fill-in-the-Middle (FIM) tasks---plausible but incorrect completions such as invented API methods, invalid parameters, undefined variables, or non-existent imports.
These failures pass superficial review yet introduce runtime errors.
We introduce \benchmark, a verified multi-lingual benchmark of \textbf{1,951} FIM samples across \textbf{7} languages and \textbf{4} hallucination types.
Samples are curated through an adversarial pipeline: a frontier LLM generates plausible hallucinations, four diverse judge models evaluate them, embedding-based clustering mines progressively harder examples, self-contained Docker containers verify that golden completions compile while hallucinated variants produce the expected runtime error, and a final human-expert review removes any remaining biased or trivially decidable samples.
We evaluate \textbf{11 open-weight FIM models} from five families spanning $0.5$B--$32$B parameters: a six-point Qwen2.5-Coder scaling slate, plus a cross-family slate (CodeLlama, DeepSeek-Coder-V2, StarCoder2). The strongest model reaches only $84.5\%$ pass@1, no family exceeds $0.77$ Edit Similarity, and every family produces hallucination-aligned completions on a non-trivial share of samples, confirming that the difficulty exposed by \benchmark is task-intrinsic rather than family-specific. We release the benchmark, containers, and evaluation framework at \url{https://github.com/microsoft/delulu}.
\end{abstract}

\input{sections/introduction}
\input{sections/design}
\input{sections/statistics}

\input{sections/evaluation}

\input{sections/conclusion}

\clearpage
\bibliography{references}
\bibliographystyle{plainnat}

\clearpage
\appendix
\input{appendix/additional_results}

\end{document}

%% file: sections/introduction.tex
\savespace[1]
\section{Introduction}
\label{sec:intro}
\savespace[0.6]

Large Language Models (LLMs) now drive AI code assistants used by millions of developers daily. Among completion paradigms, Fill-in-the-Middle (FIM)~\citep{bavarian2022efficient}, which generates the missing \emph{middle} given a code prefix and suffix, has become one of the dominant approaches in production systems such as GitHub Copilot~\citep{github_copilot_2025} and Cursor~\citep{cursor_2025}.
Yet FIM models routinely produce \emph{hallucinations}\footnote{``Hallucination'' has multiple definitions in the literature, ranging from any factually incorrect generation~\citep{ji2023survey_hallucination} to broader notions of unfaithfulness; in the code setting, taxonomies span logic, specification, and knowledge errors~\citep{lee2025hallucination,agarwal2025codemirage}. We use the term in the \emph{code} sense---completions that are syntactically valid and semantically plausible yet reference or rely on facts that are not true of the resolved program context.}: completions that are syntactically valid and semantically plausible but factually wrong.
A hallucinated method call like \texttt{df.remove\_nulls()} reads naturally in a pandas context but raises an \texttt{AttributeError} at runtime; a hallucinated import \texttt{from sklearn.neural import DeepClassifier} extends a real namespace with a non-existent module.
These failures pass superficial code review and propagate through dependency chains, introducing latent bugs that only manifest under specific runtime conditions~\citep{lee2025hallucination}.

Two complementary questions determine whether this problem is under control:
\begin{tcolorbox}[colback=gray!5, colframe=gray!60, boxrule=0.4pt, arc=1.5pt, left=4pt, right=4pt, top=3pt, bottom=3pt, boxsep=0pt]
\emph{Q1 (Generation):} Can FIM models reliably produce non-hallucinated completions across diverse languages and hallucination types?\\[2pt]
\emph{Q2 (Detection):} Can frontier LLMs reliably detect hallucinations when reviewing code completions?
\end{tcolorbox}
\noindent If both questions could be answered positively (that models reliably avoid generating hallucinations, and that any remaining hallucinations are caught before reaching production) the problem would be effectively solved. However, the current state of the art falls short on both fronts.

\begin{wrapfigure}{r}{0.4\linewidth}
\vspace{-10pt}
\centering
\includegraphics[width=\linewidth]{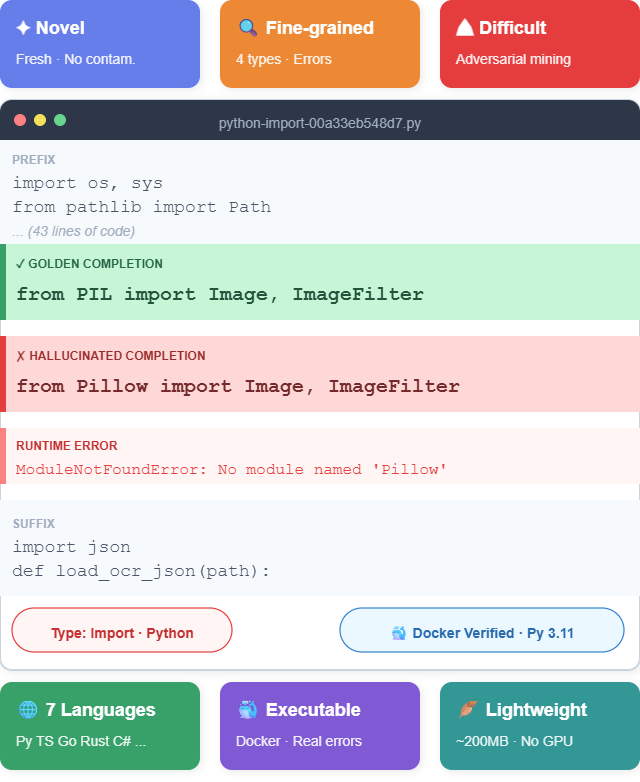}
\caption{A \benchmark sample with golden and hallucinated completions, Docker-verified across 7 languages and 4 hallucination types.}
\label{fig:delulu_overview}
\vspace{-10pt}
\end{wrapfigure}

Regarding \emph{generation}, recent models have made substantial progress on established code benchmarks: top performers now exceed $90\%$ pass@1 on HumanEval~\citep{chenEvaluatingLargeLanguage2021} and achieve competitive scores on SAFIM~\citep{gong2024safim}. Yet this progress is increasingly difficult to interpret. Models are suspected of \textit{overfitting} to the fixed problem sets of popular benchmarks~\citep{jainLiveCodeBenchHolisticContamination2024}, and the dominant FIM benchmarks (HumanEval Infilling~\citep{bavarian2022efficient}, SAFIM~\citep{gong2024safim}) are mostly Python-only, lack execution verification of the infilling task itself, and do not distinguish between different \emph{types} of errors. A model that passes HumanEval Infilling may still routinely hallucinate non-existent imports or fabricate API methods when deployed on real-world code, and no current benchmark is designed to measure this.

Regarding \emph{detection}, we ask whether a frontier LLM, acting as a code reviewer, can distinguish a correct FIM completion from a hallucinated one. To test this, we generate $\sim$40K FIM (prefix, suffix, completion) triples from public GitHub repositories and use \claude-4.5 to produce a hallucinated variant for each golden completion across four hallucination types (method, parameter, undefined variable, import, See Table~\ref{tab:types}). We then present each completion to four frontier judge models (\gptfiveone, \gptfivetwo, \gptfivetwocodex, \glm) and ask them to accept or reject it. We measure \emph{both-correct accuracy}: the fraction of samples on which the judge correctly accepts the golden completion \emph{and} correctly rejects the hallucinated one. As Table~\ref{tab:judge_iter} (Iter~0) shows, even the best judge (\gptfivetwocodex) achieves only $56$--$90\%$ both-correct accuracy,
with import hallucinations fooling all models at rates exceeding $40\%$. These numbers are measured against LLM-generated labels that have not yet been execution-verified; they therefore represent a conservative lower bound on true judge capability, since some ``hallucinated'' labels may themselves be incorrect (Appendix~\ref{app:detection_study}). Even so, the results confirm that hallucination detection in code remains an open problem for the most capable models available today.

Together, these findings reveal a compounding gap: frontier models cannot reliably \emph{detect} hallucinations, and we cannot reliably \emph{measure} whether models generate them. What is missing is a \textit{novel} benchmark that simultaneously provides (1)~systematic hallucination categorization with fine-grained types, (2)~multi-lingual coverage beyond Python, (3)~execution-based verification that hallucinations produce real errors, and (4)~adversarial difficulty calibration ensuring the benchmark is genuinely challenging.

We introduce \benchmark\footnote{The name is a play on \emph{delulu}, internet slang for ``delusional'', chosen to reflect the central property of the failure mode this benchmark targets: hallucinated completions are syntactically confident, semantically plausible, and persuasively wrong.} to address these requirements.
\benchmark is a verified multi-lingual benchmark of $1{,}951$ FIM samples across $7$ programming languages and a 4-category hallucination taxonomy (method, parameter, undefined variable, import), each producing a distinct verifiable runtime error (Figure~\ref{fig:delulu_overview}). Samples are curated through a five-stage \emph{adversarial pipeline}: hallucination generation by \claude\footnote{We piloted three candidate generators (\claude-4.5, \gptfiveone, \gptfivetwo) and found \claude exhibited the strongest prompt adherence, obeying the ``modify exactly one element, keep everything else identical, do not signal that the output is hallucinated'' constraint at the highest rate. This property is critical because the hallucinations must be statistically indistinguishable from real completions. We accordingly use \claude as the sole generator; the four-judge panel and the human-verification pass below address the residual single-source bias.}, multi-model judging by four diverse LLMs, embedding-based clustering of judge failures to surface harder examples, difficulty-based selection with language balancing, and Docker-based execution verification. To rule out single-generator bias, every finalized sample is additionally inspected by human experts and any biased or trivially decidable rows are removed before release.

We then conduct what is, to our knowledge, the broadest open-weight FIM evaluation on a hallucination-targeted benchmark to date: $11$ models from $5$ families spanning $0.5$B to $32$B parameters. The Qwen2.5-Coder-Instruct~\citep{hui2024qwen25coder} scaling slate (six models) achieves a maximum of $84.5\%$ pass@1 at $32$B, still failing on roughly one in seven unique FIM contexts. A cross-family slate of CodeLlama~\citep{roziere2023codellama}, DeepSeek-Coder-V2-Lite-Instruct~\citep{deepseekcoderv2}, and StarCoder2~\citep{lozhkov2024starcoder2} confirms that the difficulty is family-invariant: no model exceeds $0.77$ Edit Similarity, every family produces hallucination-aligned completions on $0.7$--$2.0\%$ of samples, and the strongest cross-family model (DSCoder-V2-Lite-Instruct) still trails the best Qwen by $1.8$ pass@1 points.

\savespace[0.5]
\paragraph{Related work.}
\savespace[0.2]
Existing benchmarks fall short on the two axes \benchmark targets. \emph{Function-level} suites---HumanEval~\citep{chenEvaluatingLargeLanguage2021}, MBPP~\citep{austinProgramSynthesisLarge2021}, MultiPL-E~\citep{cassano2023multipl}, BigCodeBench~\citep{zhuoBigCodeBenchBenchmarkingCode2025}, ClassEval~\citep{duClassEvalManuallyCraftedBenchmark2023}, CoderEval~\citep{yuCoderEvalBenchmarkPragmatic2024}, and contamination-resistant variants such as LiveCodeBench~\citep{jainLiveCodeBenchHolisticContamination2024} and EvoCodeBench~\citep{Li2024EvoCodeBenchAE}---target generation from natural-language specifications, not FIM completion. The dominant \emph{FIM} benchmarks (HumanEval Infilling~\citep{bavarian2022efficient}, SAFIM~\citep{gong2024safim}, CrossCodeEval~\citep{dingCrossCodeEvalDiverseMultilingual2023}) lack execution verification of the infilled code; DevBench~\citep{aaa_devbench2025} executes but measures overall completion quality rather than hallucinations, and repository-level suites such as SWE-bench~\citep{jimenezSWEbenchCanLanguage2024,dengSWEBenchProCan2025} surface hallucinations only incidentally. Recent \emph{hallucination} work taxonomizes the failure modes~\citep{lee2025hallucination} and analyzes patterns across model families~\citep{agarwal2025codemirage} but ships no execution-verified benchmark artifact. Table~\ref{tab:comparison} positions \benchmark in this landscape; to our knowledge it is the only benchmark that simultaneously offers FIM focus, a $4$-category hallucination taxonomy, multi-lingual execution verification via per-sample Docker containers, and adversarial difficulty calibration. An extended discussion is deferred to Appendix~\ref{app:related_extended}.

\input{sections/related_works}

\savespace[0.5]
\paragraph{Contributions.}
\savespace[0.2]
\begin{enumerate}[leftmargin=*,nosep]
    \item A four-category FIM hallucination taxonomy in which every category produces a distinct, verifiable runtime error, enabling execution-based ground truth (\S\ref{sec:taxonomy}).
    \item A five-stage curation pipeline that pairs each completion mined from real GitHub code with a hallucinated counterpart, filters trivially-decidable cases by repeatedly probing frontier judges, and retains only samples that compile as golden and provably fail as hallucinated (\S\ref{sec:pipeline}).
    \item $1{,}951$ execution-verified samples across 7 languages, each packaged as a self-contained Docker container with three invocation modes (\S\ref{sec:verification}).
    \item Evaluation of $11$ open-weight FIM models from $5$ families: scaling, cross-family, per-language, and per-hallucination-type analyses providing evidence that the difficulty of \benchmark generalizes beyond a single training pipeline (\S\ref{sec:eval}).
    \item A frontier-judge detection study on the verified release: $8$ judges spanning two vendors (OpenAI and Anthropic) score paired golden/hallucinated completions on all $1{,}951$ samples, and even the strongest judge (Claude-4.5-Opus) achieves only $92.1\%$ both-correct accuracy and $83.4\%$ on Import, with smaller judges falling as low as $52.3\%$---a $40$-point capability range that establishes hallucination \emph{detection} on \benchmark as unsolved at the frontier and reusable as a benchmark for code-review and verifier models (\S\ref{sec:detection}).
\end{enumerate}

%% file: sections/related_works.tex
\begin{table}[t]
\caption{Comparison of code completion and hallucination benchmarks. \benchmark uniquely combines FIM focus, hallucination categorization, execution verification, and multi-lingual coverage.}
\label{tab:comparison}
\centering
\resizebox{\linewidth}{!}{%
\begin{tabular}{lccccccc}
\toprule
\textbf{Benchmark} & \textbf{Task} & \textbf{Exec.} & \textbf{Langs} & \textbf{Samples} & \textbf{FIM} & \textbf{Hall.} & \textbf{Adv.} \\
\midrule
HumanEval~\citep{chenEvaluatingLargeLanguage2021} & Gen. & \cmark & 1 & 164 & \xmark & \xmark & \xmark \\
MBPP~\citep{austinProgramSynthesisLarge2021} & Gen. & \cmark & 1 & 974 & \xmark & \xmark & \xmark \\
MultiPL-E~\citep{cassano2023multipl} & Gen. & \cmark & 18 & --- & \xmark & \xmark & \xmark \\
HumanEval Infilling~\citep{bavarian2022efficient} & FIM & \xmark & 1 & 1{,}033 & \cmark & \xmark & \xmark \\
SAFIM~\citep{gong2024safim} & FIM & \xmark & 4 & 17{,}720 & \cmark & \xmark & \xmark \\
CrossCodeEval~\citep{dingCrossCodeEvalDiverseMultilingual2023} & FIM & \xmark & 4 & $\sim$10k & \cmark & \xmark & \xmark \\
DevBench~\citep{aaa_devbench2025} & FIM & \cmark & 6 & 1,800 & \cmark & \xmark & \xmark \\
CodeMirage~\citep{agarwal2025codemirage} & Analysis & \xmark & 1 & 1{,}137 & \xmark & \cmark & \xmark \\
SWE-bench~\citep{jimenezSWEbenchCanLanguage2024} & SWE & \cmark & 1 & 2,294 & \xmark & \xmark & \xmark \\
\midrule
\textbf{\benchmark} & \textbf{FIM} & \textbf{\cmark} & \textbf{7} & \textbf{1,951} & \textbf{\cmark} & \textbf{\cmark\,(4)} & \textbf{\cmark} \\
\bottomrule
\end{tabular}%
}
\savespace[1]
\end{table}

%% file: sections/design.tex
\savespace[0.8]
\section{Benchmark Design}
\label{sec:design}
\savespace[0.5]

\begin{figure}[t]
\centering
\includegraphics[width=0.95\linewidth]{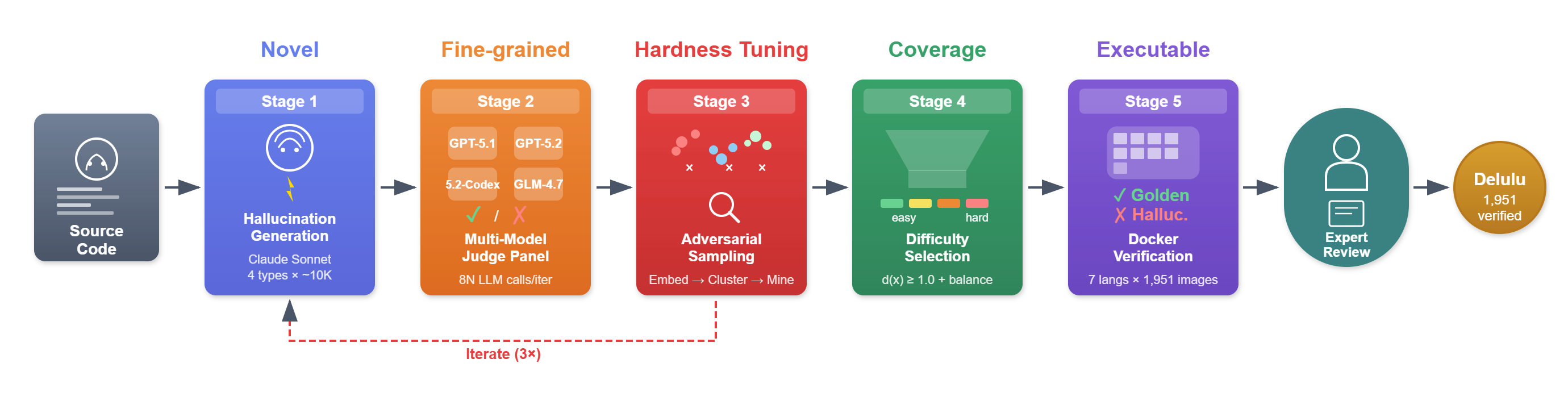}
\savespace[0.7]
\caption{The five-stage \benchmark curation pipeline.}
\savespace[1.2]
\label{fig:pipeline}
\end{figure}

\benchmark is built through a five-stage pipeline (Figure~\ref{fig:pipeline}): hallucination generation, multi-model judge evaluation, iterative adversarial sampling, difficulty-based selection, and Docker-based execution verification, followed by a human-expert review pass. We summarize the design here; full prompts, judge configurations, distributed-execution details, and the human-review protocol are in Appendix~\ref{app:pipeline_details}.

\savespace[0.6]
\subsection{Hallucination Taxonomy}
\label{sec:taxonomy}
\savespace[0.4]

\benchmark targets four categories of FIM hallucinations, each defined by a distinct semantic capability and a distinct runtime error class (Table~\ref{tab:types}): \textbf{Method} (invented method names $\Rightarrow$ \texttt{AttributeError}), \textbf{Parameter} (non-existent keyword arguments $\Rightarrow$ \texttt{TypeError}), \textbf{Undefined Variable} (out-of-scope references $\Rightarrow$ \texttt{NameError}), and \textbf{Import} (non-existent modules $\Rightarrow$ \texttt{ImportError}). Every hallucination is constrained to be \emph{plausible}---following the naming conventions of the surrounding API rather than appearing obviously synthetic---so that detection cannot rely on surface cues.

The four categories are a deliberate subset of the refined code-hallucination taxonomy of \citet{gao2025hallucination_survey}, which catalogues API Misuse, Library Misuse, and Dependency Hallucination as the primary knowledge-level failure modes in code LLMs. We retain exactly the categories that (i)~occur frequently in FIM completion, (ii)~map to a small, well-defined set of runtime error classes, and (iii)~can therefore be automatically verified by executing the assembled file. Other hallucination modes documented in the literature---logic-flow violations, deprecated-API usage, off-by-one errors, behavior-correct-but-API-wrong completions, and broader functional misalignment~\citep{lee2025hallucination, gao2025hallucination_survey}---are excluded by design: their ground truth depends on project-specific test suites, semantic-equivalence judges, or human raters, each of which would reintroduce the label noise that the execution gate is meant to eliminate. \benchmark should accordingly be read as a \emph{lower bound} on hallucination prevalence: failure on a \benchmark sample implies a production-relevant error, but passing \benchmark does not preclude semantic hallucinations of the excluded types. Extending coverage to non-error categories is the principal item on our roadmap (Appendix~\ref{app:discussion}).

The four categories share a single-element-modification protocol: the golden and hallucinated tuples differ in exactly one method name, one keyword argument, one identifier, or one import path; prefix, suffix, surrounding identifiers, and formatting are all identical. Figure~\ref{fig:delulu_overview} shows a representative Import sample illustrating this protocol; one example per category is reproduced in Appendix~\ref{app:examples}.

\savespace[0.6]
\subsection{Curation Pipeline}
\label{sec:pipeline}
\savespace[0.4]

The pipeline's goal is simple: produce a large pool of (golden, hallucinated) FIM pairs, then keep only those that are \emph{both} non-trivial---hard enough that frontier code reviewers cannot dismiss them at a glance---\emph{and} executable, so the hallucinated variant provably fails at runtime. To achieve this, we alternate three operations across iterations: (i)~generate candidate pairs from real GitHub source, (ii)~probe the candidates with a panel of frontier judges to surface failure modes, and (iii)~mine new candidates that resemble those failure modes. Each iteration thus tightens the difficulty distribution while leaving easy or already-saturated cases behind. The remainder of this section instantiates these three operations.

Starting from source files scraped from public GitHub repositories, \claude generates paired golden and hallucinated completions using type-specific prompts ($N\approx10K$ samples per type per iteration); we use \claude as the sole generator because a pre-experiment found it had the strongest prompt adherence among candidate frontier models (see footnote in \S\ref{sec:intro}). Four judges (\gptfiveone, \gptfivetwo, \gptfivetwocodex, \glm) then independently score each (sample, completion) pair as correct/incorrect with chain-of-thought reasoning, yielding $8N$ LLM calls per iteration. We analyze judge disagreements, embed their reasoning with all-MiniLM-L6-v2, cluster via K-means with silhouette-based $k$ selection, and mine semantically similar code from the unlabeled corpus (cosine similarity in $[0.75, 0.99)$). This loop runs for three iterations.

The mining loop is validated by Table~\ref{tab:judge_iter}: judge accuracy decreases across iterations, confirming that progressively harder samples are surfaced. \glm degrades most sharply on import (0.17$\to$0.04$\to$0.03), reflecting that open-weights models are particularly vulnerable to ecosystem-specific hallucinations; \gptfivetwocodex maintains $\geq 0.89$ on undefined variable, reflecting the relative simplicity of scope reasoning.
We note that some GPT-family judges show non-monotonic import accuracy across iterations (e.g., \gptfiveone: $0.52 \to 0.40 \to 0.55$). This does not contradict the overall trend: the adversarial loop mines qualitatively \emph{different} failure modes in each iteration rather than monotonically scaling a single difficulty axis. In Iteration~2, the mined import samples happen to overlap less with the failure patterns of some GPT judges, causing a partial recovery. The key validation signal is the \emph{cross-judge average}, which decreases monotonically when pooled across all four judges.

\begin{table}[t]
\centering
\begin{minipage}[t]{0.38\linewidth}
\caption{Hallucination taxonomy. Each type produces a distinct, verifiable runtime error.}
\label{tab:types}
\centering\small
\resizebox{\linewidth}{!}{%
\begin{tabular}{lll}
\toprule
\textbf{Type} & \textbf{Error Class} & \textbf{Semantic Req.} \\
\midrule
Method & AttributeError & API method knowledge \\
Parameter & TypeError & Signature knowledge \\
Undef.\ Var. & NameError & Scope tracking \\
Import & ImportError & Package ecosystem \\
\bottomrule
\end{tabular}%
}
\end{minipage}%
\hfill
\begin{minipage}[t]{0.60\linewidth}
\caption{Both-correct accuracy across adversarial iterations (Iter 0 / Iter 1 / Iter 2). Decreasing accuracy in most cases confirms harder samples are mined over time.}
\label{tab:judge_iter}
\centering\small
\resizebox{\linewidth}{!}{%
\begin{tabular}{lcccc}
\toprule
\textbf{Judge} & \textbf{Import} & \textbf{Method} & \textbf{Parameter} & \textbf{Undef.\ Var.} \\
\midrule
\glm & .17\,/\,.04\,/\,.03 & .31\,/\,.16\,/\,.22 & .23\,/\,.17\,/\,.28 & .42\,/\,.21\,/\,.50 \\
\gptfiveone & .52\,/\,.40\,/\,.55 & .64\,/\,.55\,/\,.52 & .57\,/\,.58\,/\,.56 & .78\,/\,.76\,/\,.77 \\
\gptfivetwo & .52\,/\,.43\,/\,.54 & .62\,/\,.65\,/\,.52 & .56\,/\,.55\,/\,.56 & .74\,/\,.71\,/\,.72 \\
\gptfivetwocodex & .56\,/\,.48\,/\,.59 & .77\,/\,.76\,/\,.67 & .65\,/\,.64\,/\,.64 & .90\,/\,.89\,/\,.89 \\
\bottomrule
\end{tabular}%
}
\end{minipage}
\savespace[1]
\end{table}

\savespace[0.6]
\paragraph{Difficulty selection.}
We score each sample by $d(x) = \sum_{j} \mathbb{1}[\text{judge}_j(x_{\text{hall}})=1] + 0.5\sum_{j} \mathbb{1}[\text{judge}_j(x_{\text{gold}})=0]$ and require $d(x) \geq 1.0$. This equation essentially filters out samples that could not fool enough models. Language balancing targets $\lfloor N/7 \rfloor$ samples per language.

\savespace[0.6]
\subsection{Execution Verification}
\label{sec:verification}
\savespace[0.5]

Every released sample is \textbf{execution-verified}: ground-truth labels do not rely on heuristics or LLM judgments. For each sample we assemble two complete source files (prefix + completion + suffix) and run them inside a per-language Docker container. The golden file must compile/parse and run without errors; the hallucinated file must produce the \emph{specific} error class corresponding to its hallucination category (\texttt{AttributeError}, \texttt{TypeError}, \texttt{NameError}, or \texttt{ImportError}). Samples failing either condition are discarded.
Verification rests on three components, summarized below; full details are in Appendix~\ref{app:docker_full}.

\begin{itemize}[leftmargin=*,itemsep=2pt,topsep=2pt]
\item \textbf{Per-language base images.} We maintain seven base images covering Python (CPython 3.11), Java (OpenJDK 17), TypeScript (Node.js 20), Go (1.21), Rust (nightly), C\# (.NET 8.0), and C++ (GCC 13/Clang). Each ships with the standard package manager plus a dependency-resolution module that installs the source file's imports before verification.
\item \textbf{Standalone-file precheck.} Each \benchmark sample is a single source file mined from a large open-source project, and most files in such projects depend transitively on sibling files in the same repository. To keep every per-sample container lightweight and self-contained, a static precheck discards files whose dependency graph is not closable within a single file. The precheck retains roughly $5\%$ of mined files and is the dominant determinant of overall yield.
\item \textbf{Docker verification with a \claude-assisted fix loop.} Conditional on passing the precheck, $\sim\!0.4$ of the (golden, hallucinated) tuples verify on the first attempt. For the remainder, a \claude-assisted fix loop (up to three iterations) inspects the failure trace and proposes patches to the container environment---installing missing system libraries, adding package dependencies, or adjusting build flags---raising the verification rate to $\sim\!0.75$ on standalone files. Successfully verified samples are baked into self-contained Docker containers in Dockerhub exposing three modes: \texttt{verify golden}, \texttt{verify hallucinated}, and \texttt{verify patch} (which evaluates a model-generated completion supplied via stdin), enabling drop-in integration with any FIM evaluation pipeline.
\end{itemize}

\savespace[0.6]
\subsection{Human Expert Review}
\label{sec:human_review}
\savespace[0.4]

After execution verification, every Docker-verified sample is reviewed by a panel of three human experts via a dedicated annotation interface (Figure~\ref{fig:expert_panel} in Appendix~\ref{app:human_review}). The interface shows reviewers the prefix, suffix, and \emph{both} completions \emph{labelled} as golden and hallucinated---so the task is not to guess which is which but to validate that the labels are correct and the pair is non-trivially distinguishable. Each expert then selects one of three actions: \emph{accept} (the tuple is valid as-is), \emph{reject} (the tuple is fundamentally flawed), or \emph{edit} (the hallucinated completion is modified to better fit the context). Edited samples are re-run through the Docker execution pipeline to confirm that the revised hallucinated completion still triggers the expected runtime error. Of $1{,}957$ Docker-verified candidates, $1{,}744$ were accepted as-is, $6$ were rejected, and $207$ were edited and re-verified, yielding the final $1{,}951$ released samples. This post-verification human pass is what guarantees the benchmark is free from residual generator bias and label noise.

%% file: sections/statistics.tex
\savespace[0.8]
\section{Benchmark Statistics}
\label{sec:stats}
\savespace[0.6]

\paragraph{Distribution.}
The final \benchmark dataset contains $1{,}951$ execution-verified samples distributed across seven programming languages and all four hallucination types; because some FIM contexts share a prefix, suffix, and golden completion across more than one hallucinated variant, these $1{,}951$ samples cover $950$ unique FIM contexts. Table~\ref{tab:cross} shows the joint distribution. TypeScript contributes the most samples ($420$, $21.5\%$), followed by Python ($374$, $19.2\%$) and Go ($291$, $14.9\%$); undefined variable is the most frequent category ($577$) and parameter the least ($435$).

Two cells stand out and are expected rather than incidental. Python$\times$Parameter is small ($22$) because dynamically typed Python functions often accept arbitrary keyword arguments without raising \texttt{TypeError}, so very few mined Python sites admit a clean parameter hallucination. C++ contributes the fewest samples per language ($125$) because the standalone-file precheck (\S\ref{sec:verification}) retains far fewer C++ files than other languages: most C++ source files in real repositories rely on headers and translation units that live in sibling files. As a general rule, per-(language, type) cells below $\sim\!50$ samples should be read as suggestive rather than confirmatory. To address the imbalance directly, we are actively curating an extended release in which every cell contains at least $100$ verified samples; the present paper releases the current $1{,}951$-sample artifact and reports the extended-release timeline on the project page.

\savespace[0.6]
\paragraph{Provenance and Freshness.}
Source code for the benchmark is drawn from a code-mining pipeline that extracts API call sites from public GitHub repositories. The pipeline targets ${\sim}25$ third-party packages per language, selected by adoption frequency in anonymized code completion telemetry, LLM-based ecosystem ranking, and expert endorsement. For languages with standardized package managers, only repositories declaring a dependency on a target package version released within 4 months of the mining date are retained. For the current version, mined in October 2025, target package versions span April--September 2025, meaning the source repositories had adopted recent library releases at the time of collection. C++ libraries are selected via expert curation rather than version constraints. The 1{,}951 pre-hallucination examples span 319 repositories (median 81 GitHub stars), 510 source files, and 7 languages.

\savespace[0.6]
\paragraph{Complexity.}
Beyond the hallucination axis, a benchmark's difficulty also depends on the structural complexity of the code surrounding each completion: a model that handles short, self-contained snippets may still fail once the surrounding code grows longer or its control flow becomes more intricate. We therefore characterize each \benchmark sample along four \emph{code-complexity} dimensions and compare \benchmark to existing FIM benchmarks on the same axes (Table~\ref{tab:complexity}): \emph{prefix lines of code} (Pfx LOC), the amount of context the model reads before the hole; \emph{total LOC} (Tot LOC), the length of the fully assembled source file; \emph{completion tokens} (Cmp Tok.), the output length measured via \texttt{cl100k\_base}; and \emph{cyclomatic complexity} (CC)~\citep{landman2016empirical}, the number of linearly independent paths through the code, which proxies the control-flow reasoning required to fill the hole.

We can observe in Table~\ref{tab:complexity} that \benchmark has the highest cyclomatic complexity among all compared benchmarks ($16.4$, roughly $2\times$ SAFIM and $3\times$ DevBench), because samples are drawn from real-world GitHub repositories with non-trivial branching, looping, and error handling. It also has the highest total file length among benchmarks whose completions span more than one line ($152.6$ LOC vs.\ $65.3$ for DevBench), meaning models must reason over substantially more surrounding context. Per-language complexity varies: C\# samples average $255.3$ LOC and $22.1$ CC (enterprise-style code), while TypeScript averages $77.0$ LOC and $9.6$ CC (compact web-development patterns). The full per-language breakdown is in Appendix~\ref{app:stats_full}.

\begin{table}[t]
\centering
\begin{minipage}[t]{0.35\linewidth}
\caption{Sample distribution: language $\times$ hallucination type.}
\label{tab:cross}
\centering\small
\resizebox{\linewidth}{!}{%
\begin{tabular}{lrrrr|r}
\toprule
\textbf{Lang.} & \textbf{Imp.} & \textbf{Meth.} & \textbf{Par.} & \textbf{Und.} & \textbf{Tot.} \\
\midrule
C++ & 24 & 32 & 35 & 34 & 125 \\
C\# & 26 & 51 & 82 & 87 & 246 \\
Go & 50 & 81 & 79 & 81 & 291 \\
Java & 44 & 59 & 65 & 75 & 243 \\
Python & 166 & 78 & 22 & 108 & 374 \\
Rust & 57 & 60 & 66 & 69 & 252 \\
TS & 111 & 100 & 86 & 123 & 420 \\
\midrule
\textbf{Total} & 478 & 461 & 435 & 577 & \textbf{1951} \\
\bottomrule
\end{tabular}%
}
\end{minipage}%
\hfill
\begin{minipage}[t]{0.62\linewidth}
\caption{Complexity comparison \\across FIM benchmarks. }
\label{tab:complexity}
\centering\small
\resizebox{\linewidth}{!}{%
\begin{tabular}{lrrrrr}
\toprule
\textbf{Benchmark} & \textbf{N} & \textbf{Pfx LOC} & \textbf{Tot LOC} & \textbf{Cmp Tok.} & \textbf{CC} \\
\midrule
HumanEval Inf. & 409 & 20.1 & 21.8 & 17.4 & 4.2 \\
DevBench & 1,800 & 41.4 & 65.3 & 44.2 & 5.5 \\
SAFIM & 22,291 & 65.4 & 66.3 & 18.2 & 8.1 \\
Real-FIM-Eval & 17,879 & 222.1 & 222.1 & --- & 6.1 \\
\midrule
\textbf{\benchmark} & \textbf{1,951} & \textbf{61.7} & \textbf{152.6} & \textbf{15.8} & \textbf{16.4} \\
\bottomrule
\end{tabular}%
}
\end{minipage}
\end{table}

%% file: sections/evaluation.tex
\savespace[0.8]
\section{Evaluation}
\label{sec:eval}
\savespace[0.5]

\paragraph{Models.}
We evaluate two open-weight FIM slates totaling $11$ models from $5$ families. The \emph{primary slate} is the six-point Qwen2.5-Coder-Instruct family~\citep{hui2024qwen25coder} ($0.5$B, $1.5$B, $3$B, $7$B, $14$B, $32$B), chosen for its standardized FIM tokens and broad parameter range. The \emph{cross-family slate} adds five FIM-capable models from four independent training pipelines: CodeLlama-7B/13B~\citep{roziere2023codellama}, DeepSeek-Coder-V2-Lite-Instruct (16B MoE)~\citep{deepseekcoderv2}, and StarCoder2-7B/15B~\citep{lozhkov2024starcoder2}.

\savespace[0.5]
\paragraph{Setup.}
All models use greedy decoding (temperature $=0$) with a maximum output length of $256$ tokens, run on Azure ML batch endpoints over the full $1{,}951$ samples. For base FIM models without an instruction-tuned stop boundary (StarCoder2, CodeLlama), we apply the line-count truncation convention used by HumanEval-Infilling~\citep{bavarian2022efficient} and SAFIM~\citep{gong2024safim}, truncating each prediction to the gold completion's line count before scoring. The smallest instruct model (Qwen2.5-Coder-0.5B-Instruct) and base StarCoder2 frequently emit degenerate continuations (e.g. repeated boilerplate, or runs that never emit an end-of-completion marker) so we additionally cap their decoding at the gold completion's line budget; this prevents a single degenerate generation from cascading into spurious compile errors that would otherwise be charged against the model. Because some FIM contexts are paired with multiple hallucination variants during curation, the $1{,}951$ samples cover $950$ unique (prefix, suffix, golden) tuples, and a model only sees the FIM context---identical predictions are emitted for all siblings of a unique context. We therefore report \textbf{pass@1 over the $950$ unique contexts} (one representative per group) as the headline metric throughout this section; the per-sample view differs by less than $0.02$ absolute points on every model and the family ranking is unchanged. Alongside pass@1 we report four static metrics on every prediction: \textbf{EM} (Exact Match) is the fraction of completions identical to the gold completion byte-for-byte; \textbf{ES} (Edit Similarity) is character-level normalized Levenshtein similarity to the gold; \textbf{CB} (CodeBLEU)~\citep{ren2020codebleu} weights $n$-gram, AST, and dataflow agreement using language-specific parsers; and \textbf{HR} (Hallucination Rate) is the share of predictions whose SequenceMatcher similarity to the hallucinated variant exceeds that to the gold. Formal definitions are in Appendix~\ref{app:metrics}.  Static metrics (EM, ES, CB, HR) are reported at the per-sample level since they are computed directly on each prediction.

\savespace[0.6]
\subsection{Results}
\savespace[0.4]

\begin{wrapfigure}{r}{0.5\linewidth}
\vspace{-1.2em}
\centering
\includegraphics[width=\linewidth]{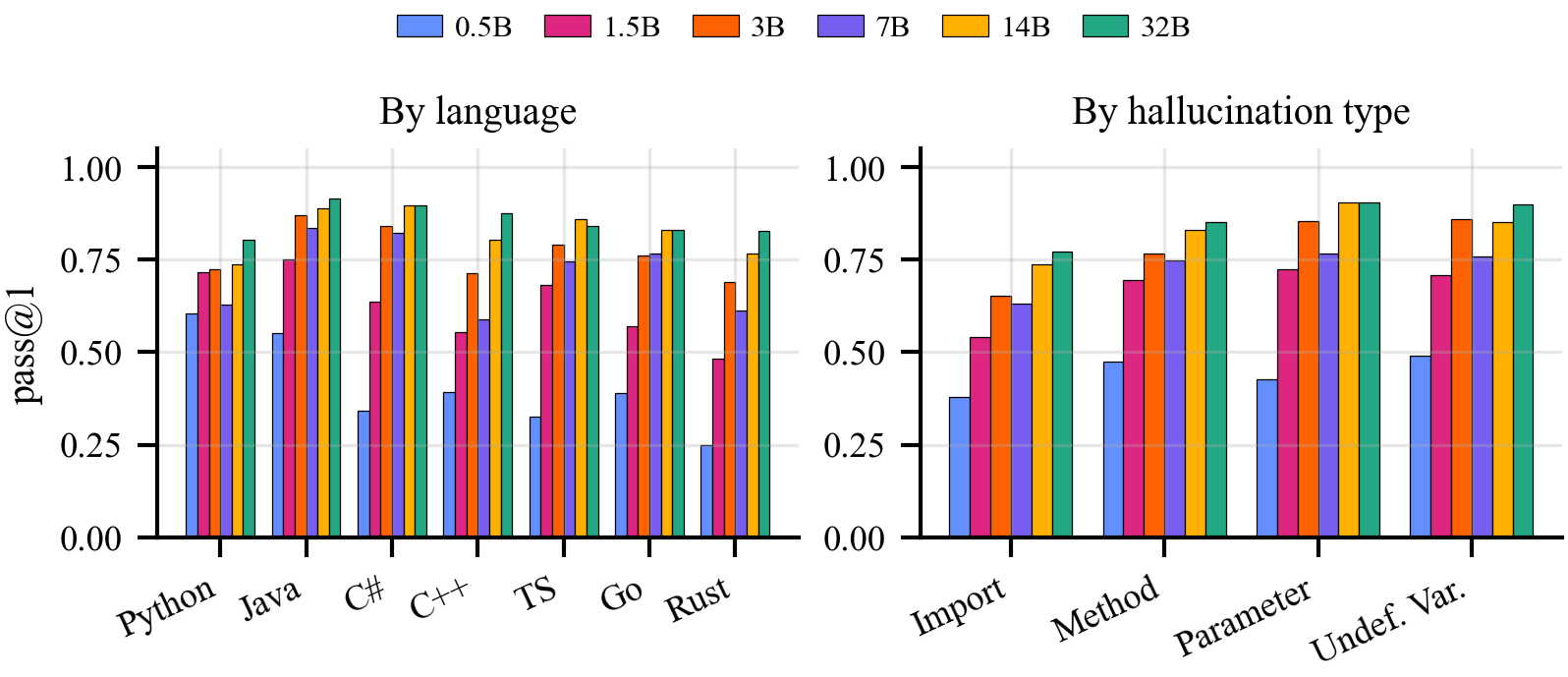}
\captionof{figure}{Qwen2.5-Coder scaling on \benchmark: pass@1 by language (left) and hallucination type (right). Import is hardest at every scale; Rust and Python are the most challenging languages.}
\label{fig:eval_overview}
\includegraphics[width=\linewidth]{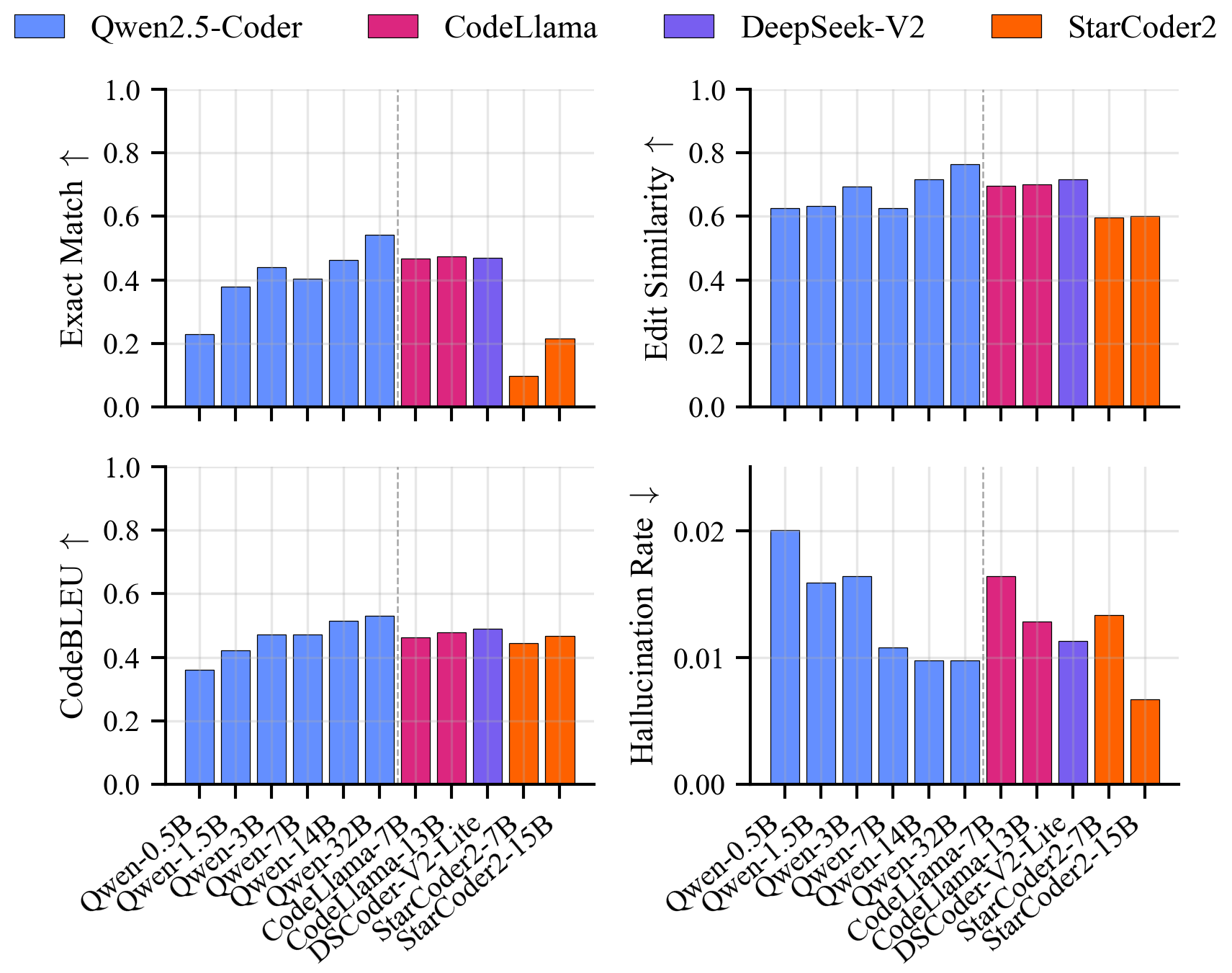}
\captionof{figure}{Cross-family static metrics on \benchmark. Dashed line separates the Qwen scaling slate (left) from the cross-family slate (right).}
\label{fig:cross_family}
\includegraphics[width=\linewidth]{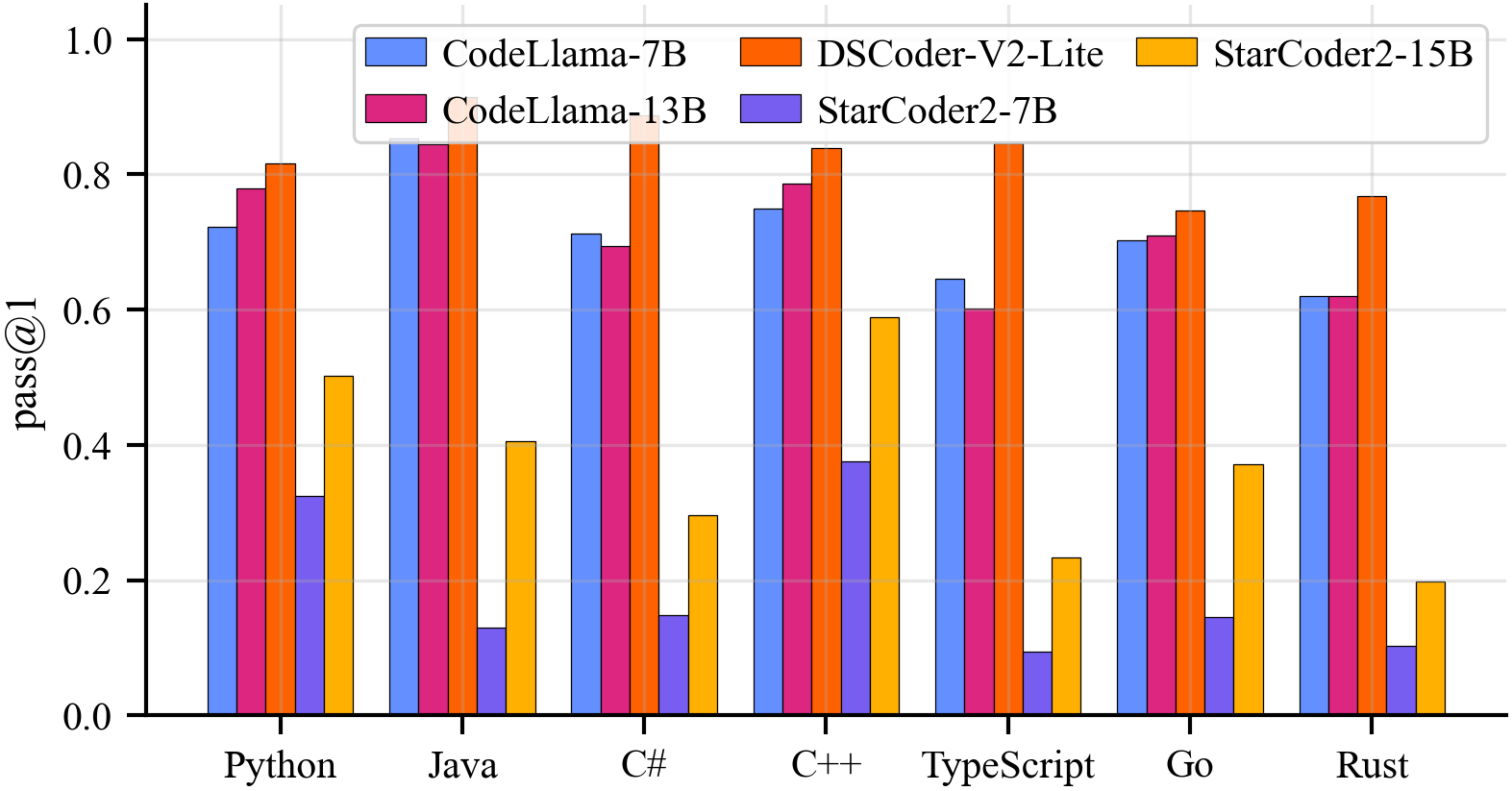}
\captionof{figure}{Per-language pass@1 for the cross-family slate.}
\label{fig:cross_family_pass1}
\vspace{-5em}
\end{wrapfigure}

Table~\ref{tab:all_results} presents the unified results for all $11$ models and Figure~\ref{fig:eval_overview} visualizes the Qwen2.5-Coder scaling trends.

\begin{table}[t]
\savespace
\caption{Results on \benchmark for all $11$ evaluated models. EM, ES, CB, and HR are computed on all $1{,}951$ samples; \textbf{pass@1} is computed on the $950$ unique (prefix, suffix, golden) FIM contexts (one representative per duplicate group) since a model emits identical predictions for siblings of the same context. Active params shown for the MoE model. \textbf{Bold}\,=\,best in column; \underline{underline}\,=\,second best.}
\label{tab:all_results}
\centering\small
\resizebox{\linewidth}{!}{%
\begin{tabular}{llrrrrrr}
\toprule
\textbf{Family} & \textbf{Model} & \textbf{Params} & \textbf{EM\,$\uparrow$} & \textbf{ES\,$\uparrow$} & \textbf{CB\,$\uparrow$} & \textbf{HR\,$\downarrow$} & \textbf{pass@1\,$\uparrow$} \\
\midrule
Qwen2.5-Coder & Qwen2.5-Coder-0.5B-Instruct & 0.5B & .228 & .625 & .360 & .020 & .436 \\
Qwen2.5-Coder & Qwen2.5-Coder-1.5B-Instruct & 1.5B & .378 & .632 & .423 & .016 & .647 \\
Qwen2.5-Coder & Qwen2.5-Coder-3B-Instruct   & 3B   & .440 & .693 & .472 & .017 & .767 \\
Qwen2.5-Coder & Qwen2.5-Coder-7B-Instruct   & 7B   & .402 & .625 & .472 & .011 & .711 \\
Qwen2.5-Coder & Qwen2.5-Coder-14B-Instruct  & 14B  & .462 & .716 & \underline{.515} & \underline{.010} & \underline{.815} \\
Qwen2.5-Coder & Qwen2.5-Coder-32B-Instruct  & 32B  & \textbf{.541} & \textbf{.765} & \textbf{.531} & \underline{.010} & \textbf{.845} \\
\midrule
CodeLlama     & CodeLlama-7B-hf              & 7B   & .468 & .695 & .463 & .017 & .711 \\
CodeLlama     & CodeLlama-13B-hf             & 13B  & \underline{.473} & .701 & .478 & .013 & .719 \\
DeepSeek-V2   & DSCoder-V2-Lite-Instruct     & 16B (2.4B*) & .469 & \underline{.717} & .489 & .011 & \underline{.827} \\
StarCoder2    & StarCoder2-7B                & 7B   & .097 & .596 & .445 & .013 & .193 \\
StarCoder2    & StarCoder2-15B               & 15B  & .215 & .600 & .466 & \textbf{.007} & .372 \\
\bottomrule
\end{tabular}%
}
\\[2pt]\footnotesize{*MoE active parameters per token.}
\savespace[1]
\end{table}

\savespace[0.6]
\paragraph{Scaling behavior (Qwen2.5-Coder).}
Four observations stand out within the primary slate. (1)~Scaling improves pass@1 from $43.6\%$ ($0.5$B) to $84.5\%$ ($32$B), a $41$-point range. (2)~At $0.5$B, text quality is already a clear bottleneck: even after the line-budget cap, only $22.8\%$ of completions match the gold byte-for-byte and Edit Similarity is $0.625$, well below every other Qwen size; $1.5$B closes most of that gap ($37.8\%$ EM, $0.632$ ES, $64.7\%$ pass@1). (3)~$3$B achieves $76.7\%$ pass@1 versus $7$B's $71.1\%$, with the $7$B model showing \emph{lower} Edit Similarity ($0.625$ vs.\ $0.693$) and Exact Match ($0.402$ vs.\ $0.440$) than $3$B; additional parameters in this range do not translate to better hallucination avoidance. (4)~Even at $32$B, $15.5\%$ of unique contexts still fail. By hallucination type, import is hardest at every scale ($77.1\%$ for $32$B vs.\ $85.1$--$90.6\%$ on the other three categories), reflecting that detecting fake imports requires memorized knowledge of the package ecosystem rather than local pattern completion. Per-language and per-language-$\times$-type breakdowns are in Figure~\ref{fig:eval_overview} and Appendix~\ref{app:fine_grained}.
The $3$B$>$$7$B inversion is reproducible across all four metrics in our single-seed run, but we do not have sufficient evidence to attribute it to a specific cause. The proximate observation is that Qwen2.5-Coder-7B-Instruct over-generates on FIM contexts where the $3$B variant stops cleanly, but identifying the underlying cause would require multi-seed verification, layer-wise probing, and access to training-recipe details. We therefore report the inversion as an artifact of a single family on this benchmark, do not generalize from it, and list a deeper investigation as an explicit open question (Appendix~\ref{app:discussion}).

\savespace[1]
\paragraph{Cross-family Comparison.}\label{sec:cross_family}
To verify that the difficulty exposed by \benchmark is intrinsic rather than an artifact of the Qwen2.5-Coder pretraining recipe, we examine the cross-family rows of Table~\ref{tab:all_results} alongside Figure~\ref{fig:cross_family}.

Four observations. (1)~\textbf{The benchmark is hard for every family we tested, not just Qwen.} If \benchmark were merely surfacing a Qwen-specific weakness, models from other training pipelines should clear it easily. They do not: across CodeLlama, DeepSeek-Coder-V2, and StarCoder2, no cross-family model exceeds $0.72$ Edit Similarity or $0.49$ CodeBLEU---essentially the same ceiling reached by the best Qwen-14B/32B checkpoints---so the difficulty appears to be a property of the task rather than of any single model lineage. 

\noindent
(2)~\textbf{Hallucination is universal.} Every model surveyed produces hallucination-aligned completions on $0.7$--$2.0\%$ of samples; StarCoder2-15B has the lowest similarity-based HR ($0.007$) yet still trails Qwen-32B's Edit Similarity by $16.5$ absolute points. 

\noindent
(3)~\textbf{Instruction tuning helps the model know \emph{when to stop}, not \emph{what to write}.} Base FIM models such as StarCoder2 frequently keep generating past the gold completion, emitting trailing tokens that are syntactically reasonable but extend beyond the intended hole; this overshoot is penalized harshly by Exact Match (which requires byte-level equality with the gold completion) but barely affects Edit Similarity and CodeBLEU, which tolerate small extra suffixes. The instruction-tuned families (Qwen-Instruct, DSCoder-V2-Lite-Instruct) therefore score much higher on Exact Match than StarCoder2 yet differ by less than three absolute points on Edit Similarity and CodeBLEU. The capability gap that instruction tuning closes is boundary detection (where the completion ends), not the underlying knowledge needed to avoid hallucinating an identifier in the first place---which is precisely what \benchmark targets. 

\noindent
(4)~\textbf{pass@1 spans a wide range.} DSCoder-V2-Lite-Instruct is the strongest cross-family model ($0.827$); CodeLlama-7B/13B cluster at $0.711$--$0.719$; base StarCoder2 drops to $0.19$--$0.37$. Even the best cross-family pass@1 trails Qwen-32B's $0.845$ by $1.8$ points. Per-language pass@1 (Figure~\ref{fig:cross_family_pass1}) shows Java is uniformly easiest ($\geq 0.85$ on every instruction-tuned cross-family model) and Rust the hardest, mirroring the within-family pattern.

\savespace[0.6]
\subsection{Hallucination detection as an LLM-judge task}
\label{sec:detection}
\savespace[0.4]

Beyond measuring whether FIM models \emph{generate} hallucinations, \benchmark can also measure whether frontier models can \emph{detect} them. Given a completion and its surrounding context, can a general-purpose LLM determine whether the completion is correct? This reframes the same samples as a binary classification task and lets us test whether the difficulty ordering observed in generation (\S\ref{sec:cross_family}), where import hallucinations are hardest, persists when the task shifts from generation to verification. Unlike the Iter~0 judge evaluation used during curation (Table~\ref{tab:judge_iter}), which was measured against unverified LLM-generated labels, the experiment below evaluates judges against \benchmark's execution-verified and human-reviewed ground truth.

\savespace[0.6]
\paragraph{Protocol.}
We present each of the $1{,}951$ samples to a judge model twice: once with the golden completion inserted between the prefix and suffix, and once with the hallucinated completion. The judge uses chain-of-thought reasoning and returns a binary score ($1$ = correct, $0$ = hallucinated). A sample is \emph{both-correct} only if the judge accepts the golden completion \emph{and} rejects the hallucinated one. We report the conjunction rather than only marginals because each gold/hallucinated pair shares a single source file: a constant-bias judge achieves a high marginal on one direction but zero on the conjunction. Precision, Recall, F1, and MCC are in Appendix~\ref{app:detection_extra}; the judge prompt is in Appendix~\ref{app:judge_prompts}.

\begin{table}[t!]
\caption{LLM-as-judge hallucination detection on \benchmark ($1{,}951$ samples). Both\,=\,conjunction of golden acceptance and hallucination rejection. \textbf{Bold}\,=\,best; \underline{underline}\,=\,second.}
\label{tab:detection}
\centering\small
\resizebox{\linewidth}{!}{%
\begin{tabular}{llrrr|rrrr}
\toprule
\textbf{Vendor} & \textbf{Judge} & \textbf{Gold\,$\uparrow$} & \textbf{Hall\,$\uparrow$} & \textbf{Both\,$\uparrow$} & \textbf{Import} & \textbf{Method} & \textbf{Param.} & \textbf{Undef.} \\
\midrule
OpenAI    & GPT-4o-mini       & .791 & .716 & .523 & .487 & .546 & .461 & .580 \\
OpenAI    & GPT-4.1-mini      & .869 & .803 & .677 & .569 & .730 & .635 & .755 \\
OpenAI    & GPT-5.4-mini      & .894 & .849 & .748 & .692 & .782 & .663 & .831 \\
Anthropic & Claude-4.5-Haiku  & .941 & .836 & .779 & .656 & .771 & .743 & .915 \\
OpenAI    & GPT-5.4           & .962 & .865 & .828 & .679 & .842 & .798 & .946 \\
Anthropic & Claude-4.6-Sonnet & .988 & .870 & .858 & .803 & .784 & .885 & .943 \\
OpenAI    & GPT-5.2-Codex     & .969 & .912 & .882 & .754 & \underline{.927} & .872 & .962 \\
Anthropic & Claude-4.5-Opus   & \textbf{.982} & \textbf{.938} & \textbf{.921} & \textbf{.834} & \textbf{.946} & \textbf{.911} & \textbf{.981} \\
\bottomrule
\end{tabular}%
}
\savespace[1]
\end{table}

\savespace[0.6]
\paragraph{Results.}
Table~\ref{tab:detection} reports results for $8$ judge models spanning two vendors (OpenAI and Anthropic), selected to form comparable pairs across capability tiers. Three findings. (1)~\textbf{Even the strongest model leaves a gap.} Claude-4.5-Opus leads at $92.1\%$ both-correct, followed by GPT-5.2-Codex ($88.2\%$); no model fully solves the benchmark. Smaller models such as GPT-4o-mini ($52.3\%$) and GPT-4.1-mini ($67.7\%$) score far lower, confirming that the benchmark discriminates across a $40$-point capability range. (2)~\textbf{The difficulty ordering is vendor-invariant.} Import hallucinations are the hardest category for every model from both vendors, while undefined-variable hallucinations are consistently the easiest, mirroring the generation results from \S\ref{sec:cross_family}. Claude-4.5-Opus reaches $83.4\%$ on imports yet $98.1\%$ on undefined variables; GPT-5.2-Codex shows a similar $21$-point gap ($75.4\%$ vs.\ $96.2\%$). (3)~\textbf{The bottleneck is hallucination rejection, not golden acceptance.} Stronger judges achieve $\geq 89\%$ golden acceptance, but even Claude-4.5-Opus misses $\sim$6\% of hallucinated completions. This confirms that the package-ecosystem knowledge required to distinguish real from fake imports is a genuine capability gap that persists across model families, vendors, and task framing. Per-language detection results are in Appendix~\ref{app:detection_study}.

%% file: sections/conclusion.tex
\savespace[0.8]
\section{Discussions}
\label{sec:conclusion}
\savespace[0.3]


\paragraph{Limitations.}
\savespace[0.4]
\benchmark deliberately scopes its claims. (i)~The release contains $1{,}951$ samples ($950$ unique FIM contexts); per-(language,\,type) cells below $\sim\!50$ samples should be read as suggestive, and an \textbf{extended release} with $\geq\!100$ verified samples per cell is in curation. (ii)~Containers are restricted to single-file standalone scenarios, so multi-file and repository-level FIM contexts are out of scope. (iii)~The $4$-category taxonomy targets knowledge-level hallucinations that produce a categorical runtime error; logic, type-mismatch, and deprecated-API failures are excluded, making \benchmark a lower bound on hallucination prevalence. (iv)~The artifact is pinned to its October 2025 ecosystem snapshot and is released as test-only; fine-tuning on it is discouraged. A more detailed treatment is in Appendix~\ref{app:discussion}.

\savespace[0.3]
\paragraph{Outlook.}
\savespace[0.4]
The persistent import gap on both the strongest generators ($\sim\!23$-point pass@1 deficit at $32$B) and the strongest judges (Claude-4.5-Opus, $\sim\!16$-point recall deficit) suggests that targeted training-data curation and verifier-style fine-tuning, rather than further scaling alone, are the most promising levers. The paired golden/hallucinated structure also makes \benchmark directly reusable as a benchmark for code-review and agentic-coding verifiers; we release it at \url{https://github.com/microsoft/delulu}.

%% file: appendix/additional_results.tex

\section{Extended Related Work}
\label{app:related_extended}

\paragraph{Function-level benchmarks.}
HumanEval~\citep{chenEvaluatingLargeLanguage2021} introduced functional-correctness evaluation with 164 hand-crafted Python programming problems; MBPP~\citep{austinProgramSynthesisLarge2021} scaled this to 974 crowd-sourced problems; APPS~\citep{hendrycksMeasuringCodingChallenge2021} added competitive-programming difficulty. These suites evaluate standalone function generation from natural-language descriptions---a substantially different task from FIM completion---and are Python-only and prone to training-data contamination~\citep{jainLiveCodeBenchHolisticContamination2024}.

\paragraph{Multi-lingual benchmarks.}
MultiPL-E~\citep{cassano2023multipl} transpiles HumanEval and MBPP to 18 languages; BigCodeBench~\citep{zhuoBigCodeBenchBenchmarkingCode2025} evaluates diverse function calls across 139 libraries; ClassEval~\citep{duClassEvalManuallyCraftedBenchmark2023} tests class-level generation; CoderEval~\citep{yuCoderEvalBenchmarkPragmatic2024} focuses on pragmatic generation with real-world dependencies. All remain focused on code \emph{generation} from specifications rather than FIM \emph{completion}.

\paragraph{Contamination-resistant benchmarks.}
LiveCodeBench~\citep{jainLiveCodeBenchHolisticContamination2024} uses time-stamped competitive-programming problems; EvoCodeBench~\citep{Li2024EvoCodeBenchAE} provides evolving evaluation aligned with real repositories. \benchmark complements these via fresh-source curation and adversarial mining.

\paragraph{FIM benchmarks (extended).}
HumanEval Infilling~\citep{bavarian2022efficient} adapts HumanEval to FIM with single-line, multi-line, and random-span settings, but is Python-only and lacks execution verification of the infill. SAFIM~\citep{gong2024safim} introduces syntax-aware FIM categories (block, API, control flow) across four languages, but does not verify execution. CrossCodeEval~\citep{dingCrossCodeEvalDiverseMultilingual2023} evaluates cross-file completion in repository contexts. DevBench~\citep{aaa_devbench2025} provides telemetry-driven FIM evaluation across six languages with execution and detailed diagnostics, but measures general completion quality rather than targeting hallucinations. Repository-level suites SWE-bench~\citep{jimenezSWEbenchCanLanguage2024,dengSWEBenchProCan2025} expose hallucination-prone scenarios but cannot attribute failures to specific hallucination categories.

\paragraph{Code hallucination (extended).}
\citet{lee2025hallucination} provide a comprehensive taxonomy of hallucinations in LLM-based code generation; CodeMirage~\citep{agarwal2025codemirage} analyzes hallucination patterns across model families. Both establish that hallucinations are common but provide no execution-verified benchmark. Existing detection approaches include static-analysis post-hoc verification, LLM-as-judge evaluation~\citep{applying-rlaif}, and retrieval-augmented generation. Our work uses a panel of four diverse LLM judges during \emph{curation}---not evaluation---and validates the difficulty calibration via judge-accuracy decay across iterations.

\section{Initial Judge Evaluation (Iter~0) and Label-Noise Considerations}
\label{app:detection_study}

Table~\ref{tab:judge_iter} (Iter~0 column) reports the both-correct accuracy of four frontier judge models on the initial $\sim$40K LLM-generated FIM hallucination pairs, before any adversarial mining, execution verification, or human review.

\paragraph{Protocol.}
We sampled real-world source files from public GitHub repositories, extracted FIM (prefix, suffix, completion) triples, and used \claude to generate a hallucinated variant for each golden completion across the four hallucination types. Four judges (\gptfiveone, \gptfivetwo, \gptfivetwocodex, \glm) then independently scored each (sample, completion) pair as correct/incorrect with chain-of-thought reasoning. A sample is \emph{both-correct} only if the judge accepts the golden completion \emph{and} rejects the hallucinated one.

\paragraph{Key findings.}
(1)~Even the best judge (\gptfivetwocodex) achieves only $56$--$90\%$ both-correct accuracy, meaning it either accepts a hallucinated completion or rejects a valid one on $10$--$44\%$ of samples. (2)~\textbf{Import is hardest to detect}: every judge scores below $56\%$ on imports, confirming that distinguishing real from fake package/module names requires ecosystem knowledge that even frontier models lack. \glm drops to $17\%$. (3)~\textbf{Undefined variable is easiest}: scope violations are locally verifiable, and \gptfivetwocodex reaches $90\%$. (4)~There is a large gap between \glm ($17$--$42\%$) and the GPT-family judges ($52$--$90\%$).

\paragraph{Label-noise caveat.}
Crucially, these Iter~0 numbers are measured against \emph{LLM-generated labels that have not been execution-verified}. Since \claude's hallucination generation is imperfect, some samples labeled ``hallucinated'' may in fact be valid code (and vice versa). When a judge ``incorrectly'' accepts such a sample, it may actually be right; the noisy label penalizes it. The Iter~0 accuracy therefore represents a \emph{conservative lower bound} on true judge capability. This is precisely why Docker-based execution verification (Stage~5, \S\ref{sec:verification}) and human expert review are essential: they eliminate label noise before the benchmark is finalized. The formal detection experiment on the verified \benchmark dataset (\S\ref{sec:detection}) measures judge capability against execution-grounded labels and should be interpreted independently of these Iter~0 numbers.

\section{Detection: Precision, Recall, F1, and MCC}
\label{app:detection_extra}

The main-paper detection results (\S\ref{sec:detection}, Table~\ref{tab:detection}) report golden acceptance, hallucination rejection, and their conjunction (both-correct accuracy). Some readers may prefer single-direction summary statistics; we therefore re-express the same per-judge confusion matrices as Precision, Recall, F1, and Matthews Correlation Coefficient (MCC), treating \emph{hallucination} as the positive class.

Let $\mathrm{GA}$ denote golden acceptance and $\mathrm{HR}$ denote hallucination rejection on the $N{=}1{,}951$ paired evaluations per judge. Treating each judge's $0$ output (``reject as hallucinated'') as a positive prediction, the four confusion-matrix counts are $\mathrm{TP}{=}N\cdot\mathrm{HR}$, $\mathrm{FN}{=}N(1{-}\mathrm{HR})$, $\mathrm{FP}{=}N(1{-}\mathrm{GA})$, and $\mathrm{TN}{=}N\cdot\mathrm{GA}$, from which Precision, Recall, F1, and MCC follow directly. Table~\ref{tab:detection_extra} summarizes the resulting metrics.

\begin{table}[h]
\caption{Single-direction detection metrics for the eight judges of Table~\ref{tab:detection}, derived from their Gold Acc.\ and Hall Rej.\ rates with ``hallucination'' as the positive class. \textbf{P}: Precision; \textbf{R}: Recall ($\equiv$ Hall Rej.); \textbf{F1}: harmonic mean of P and R; \textbf{MCC}: Matthews Correlation Coefficient. \textbf{Bold} = best in column.}
\label{tab:detection_extra}
\centering\small
\begin{tabular}{llrrrr}
\toprule
\textbf{Vendor} & \textbf{Judge} & \textbf{P}\,$\uparrow$ & \textbf{R}\,$\uparrow$ & \textbf{F1}\,$\uparrow$ & \textbf{MCC}\,$\uparrow$ \\
\midrule
OpenAI    & GPT-4o-mini       & .774 & .716 & .744 & .508 \\
OpenAI    & GPT-4.1-mini      & .860 & .803 & .830 & .673 \\
OpenAI    & GPT-5.4-mini      & .889 & .849 & .869 & .744 \\
Anthropic & Claude-4.5-Haiku  & .934 & .836 & .882 & .781 \\
OpenAI    & GPT-5.4           & .958 & .865 & .909 & .831 \\
Anthropic & Claude-4.6-Sonnet & .986 & .870 & .925 & .864 \\
OpenAI    & GPT-5.2-Codex     & .967 & .912 & .939 & .882 \\
Anthropic & Claude-4.5-Opus   & \textbf{.981} & \textbf{.938} & \textbf{.959} & \textbf{.921} \\
\bottomrule
\end{tabular}
\end{table}

Three observations. (1)~\textbf{Precision is uniformly high for the stronger half of the slate} ($\geq 0.934$ from Claude-4.5-Haiku onwards): when these judges flag a completion as hallucinated, they are rarely wrong, which is the operationally important property for a verifier sitting in front of a developer. The two smallest judges (GPT-4o-mini, GPT-4.1-mini) drop to $0.77$--$0.86$, mirroring their lower Gold Acc.\ in Table~\ref{tab:detection}. (2)~\textbf{Recall is the bottleneck for every judge.} The spread on Recall ($0.716$--$0.938$) is $\sim\!2\times$ the spread on Precision among the top six judges ($0.934$--$0.986$), so improvements in detection on \benchmark will primarily come from catching the missed hallucinations rather than reducing false flags on golden completions. (3)~The ranking by F1 and MCC closely tracks the ranking by both-correct accuracy in Table~\ref{tab:detection}: Claude-4.5-Opus first on both, GPT-5.2-Codex second, Claude-4.6-Sonnet third, GPT-5.4 fourth, and the smaller judges trail in the same order. The two metric families therefore agree on judge ordering; we retain both-correct as the headline metric in the main text because it directly reflects the dual-completion structure of \benchmark, but the F1/MCC summaries are robust to readers who weight false positives and false negatives differently.

\section{Curation Pipeline Details}
\label{app:pipeline_details}

\paragraph{Stage 1: Hallucination generation.}
From source files scraped from public GitHub repositories, \claude generates paired golden and hallucinated completions using type-specific prompts (\S\ref{app:prompts}). Import hallucinations require structural rearrangement so the hallucinated import becomes the FIM completion target. We generate $\sim$10K samples per type per iteration. \claude was selected as the sole generator after a pilot of four candidate frontier models (\claude, \gptfiveone, \gptfivetwo, \glm) showed it had the strongest prompt adherence---honoring the ``modify exactly one element, keep everything else identical, do not signal that the output is hallucinated'' constraint at the highest rate. Single-generator bias is mitigated downstream by the four-judge panel and the human-verification pass.

\paragraph{Stage 2: Multi-model judge evaluation.}
Four judges---\gptfiveone, \gptfivetwo, \gptfivetwocodex, \glm---independently evaluate each sample. Each judge receives the FIM context paired with either the golden or hallucinated completion and produces a binary score with chain-of-thought reasoning, yielding $8N$ LLM calls per iteration, where $N$ is the number of candidate samples evaluated in the iteration ($4$ judges $\times$ $2$ completions per sample).

\paragraph{Stage 3: Iterative adversarial sampling.}
We analyze judge disagreements (false positives and false negatives), extract reasoning embeddings using all-MiniLM-L6-v2, cluster them via K-means with silhouette-based $k$ selection (typically $k\in[5,10]$), and mine semantically similar code from the unlabeled corpus (cosine similarity in $[0.75,0.99)$). The loop runs for three iterations. Several patterns emerge from the iteration-level results in Table~\ref{tab:judge_iter}: \glm shows severe degradation on import (0.17$\to$0.04$\to$0.03), confirming that open-weights models are particularly vulnerable to ecosystem-specific hallucinations; \gptfivetwocodex maintains high accuracy on undefined variable ($\geq 0.89$), reflecting the relative simplicity of scope reasoning; method-hallucination accuracy is non-monotonic for some judges, suggesting that adversarial mining discovers qualitatively different failure modes in later iterations rather than simply scaling difficulty.

\paragraph{Stage 4: Difficulty-based selection.}
We score each sample
\begin{equation}
d(x) \;=\; w_h \sum_{j=1}^{4} \mathbb{1}[\text{judge}_j(x_{\text{hall}})=1] \;+\; w_g \sum_{j=1}^{4} \mathbb{1}[\text{judge}_j(x_{\text{gold}})=0],
\end{equation}
with $w_h=1.0$ (false acceptance) and $w_g=0.5$ (false rejection), requiring $d(x)\geq 1.0$. Language balancing targets $\lfloor N/7 \rfloor$ samples per language, prioritizing difficult samples.

The asymmetric weights reflect the asymmetric purpose of each judge call. A judge that \emph{accepts} a hallucination ($w_h{=}1.0$) provides direct positive evidence that the hallucination is non-trivial: at least one frontier model with chain-of-thought reasoning was unable to flag the error, which is exactly the failure mode \benchmark is designed to surface. A judge that \emph{rejects} a golden completion ($w_g{=}0.5$) provides only indirect, label-confounded evidence (the rejection may track sample-quality noise such as unusual identifiers, project-specific idioms, or upstream-repo bugs rather than hallucination subtlety); we retain that signal but down-weight it by half. The threshold $d(x)\geq 1.0$ corresponds to ``at least one judge fooled, or two judges over-rejecting the golden'', which on the Iter-0 pool retains the upper $\sim\!58\%$ of samples by difficulty and removes the trivially-decidable lower tail. We confirmed empirically that thresholds in $\{1.0, 1.5, 2.0\}$ shift cell sizes but do not change the cross-language ordering of difficulty, so we adopt the most permissive setting that still excludes universally-easy samples.

\paragraph{Distributed execution.}
The full pipeline runs on Databricks Asset Bundles with up to 16 concurrent judging tasks. Each iteration involves $\sim$80,000 LLM calls ($4$ judges $\times$ $2$ completions $\times$ $\sim$10K samples), with exponential-backoff retry (up to $5$ attempts, $60$\,s max delay). Judge evaluation per iteration takes $7$--$30$ hours depending on the model; clustering and mining add $\sim$20 hours per iteration. Stage breakdown: Stage~1 (Generation) $\sim$1 hour; Stage~2 (Judging) four parallel tasks of $7$--$30$ hours; Stage~3 (Clustering) $\sim$20 hours.

\section{Human Expert Review Protocol}
\label{app:human_review}

After all candidate samples pass Docker-based execution verification (\S\ref{sec:verification}), they undergo a final human expert review. This step is performed \emph{after} execution verification so that human effort is spent only on the small set of samples that have already been confirmed to compile and produce the expected errors.

\begin{figure}[h]
\centering
\includegraphics[width=0.85\linewidth]{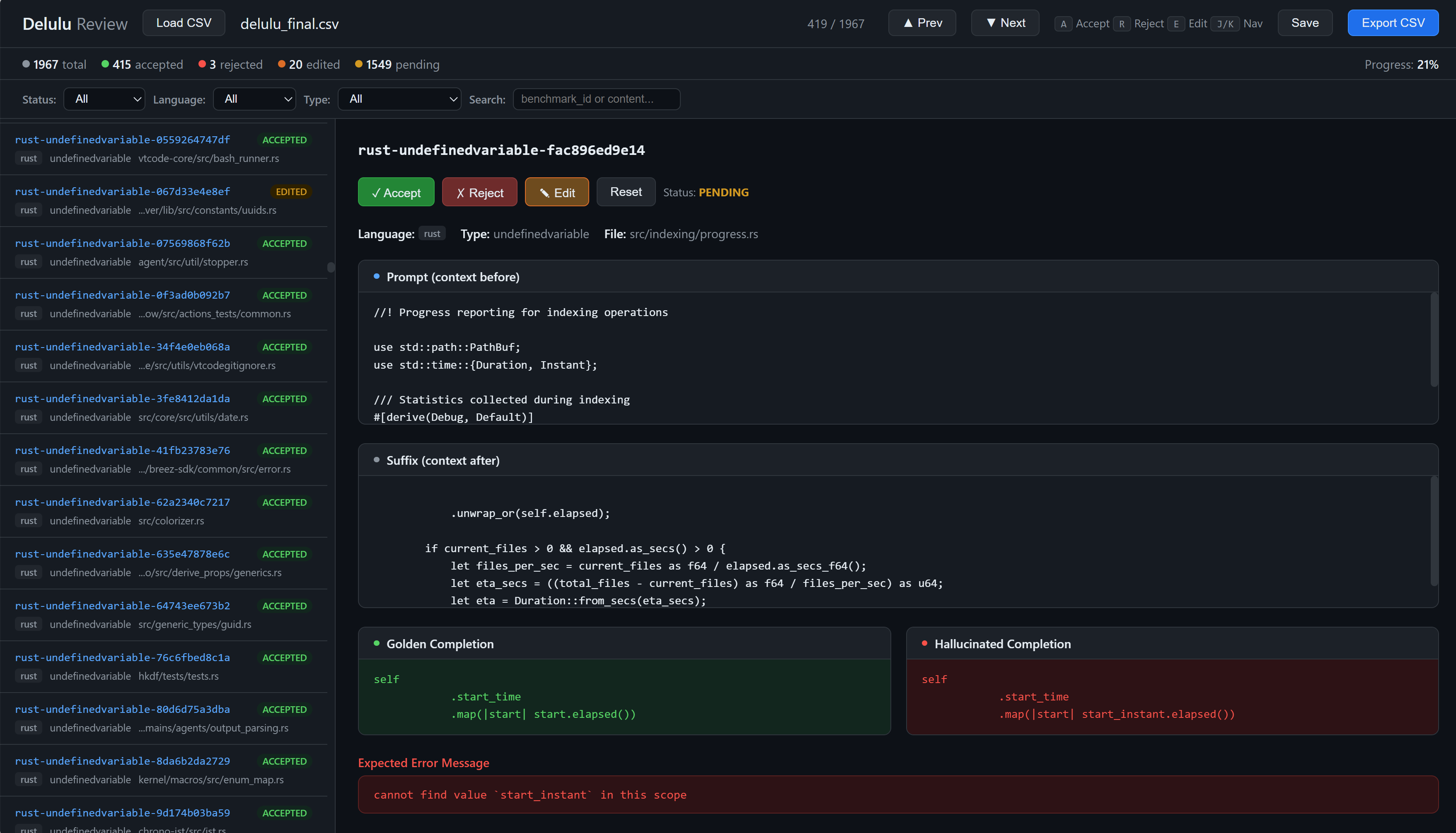}
\caption{The expert annotation interface. Each sample displays the prefix, suffix, and the two completions labelled as golden and hallucinated. Annotators can accept, reject, or edit the hallucinated completion; edits trigger automatic re-verification via Docker.}
\label{fig:expert_panel}
\end{figure}

\paragraph{Panel and workflow.}
Three expert annotators, each with professional software engineering experience across the seven target languages, review every Docker-verified sample using a dedicated annotation interface (Figure~\ref{fig:expert_panel}). The interface displays the prefix, suffix, and \emph{both} completions \emph{labelled} as golden and hallucinated; reviewers therefore validate that the labels are correct and that the pair is non-trivially distinguishable rather than guessing which is which. The annotator selects one of three actions:
\begin{itemize}[leftmargin=*,nosep]
    \item \textbf{Accept}: the tuple is valid, the golden completion is correct, and the hallucinated completion is a plausible but genuinely incorrect alternative.
    \item \textbf{Reject}: the tuple is fundamentally flawed (e.g., the golden completion is itself incorrect, or the hallucination is trivially detectable from a surface cue such as a leftover comment or formatting artifact).
    \item \textbf{Edit}: the hallucinated completion is modified by the annotator to make it more plausible or to fix a minor issue while preserving the hallucination type. Edited samples are automatically re-run through the full Docker execution pipeline to confirm that the revised hallucinated completion still triggers the expected runtime error class.
\end{itemize}

\paragraph{Statistics.}
Of $1{,}957$ Docker-verified candidate samples submitted for human review, $1{,}744$ were accepted as-is, $6$ were rejected (removed from the dataset), and $207$ received edits to the hallucinated completion (some samples were edited and subsequently accepted, so the counts overlap). All $207$ edited samples passed re-verification, confirming that the expert modifications preserved the intended runtime error. After removing the $6$ rejected samples, the final released dataset contains $1{,}951$ samples.

\section{Hallucination Generation Prompts}
\label{app:prompts}

We use type-specific system prompts for \claude. Each prompt instructs the model to modify exactly one aspect of the completion while preserving all other code. We show \textbf{method} and \textbf{import}; parameter and undefined variable follow the same structure.

\begin{promptbox}{System Prompt: Method Hallucination}
You are given a code snippet divided into three parts:
- Prefix: the portion of code before the completion.
- Completion: the part generated by a code model. The completion always includes a method call.
- Suffix: the code that comes after the completion.

Your task: simulate a hallucination by modifying the
completion:
- Replace the method name with a random, invented method name that does NOT appear in prefix/suffix.
- Keep everything else the same.
- The name should be plausible -- avoid naming it in a way that shows it is undefined or hallucinated.

Output exactly in this format:
Generated completion: <the hallucinated completion>

Do not modify prefix or suffix, add comments, or change formatting.
\end{promptbox}

\begin{promptbox}{System Prompt: Import Hallucination}
You are given a complete source code file. Read it carefully.

Your task: simulate a hallucination by REPLACING one
real import statement with a hallucinated version:
- Pick one real import line from the file.
- Replace it with a plausible but FAKE import that does NOT exist (e.g., "from datautils import "DataProcessor", "import tensorflowx").
- The fake module must NOT be a real package. It must cause an ImportError or ModuleNotFoundError.
- Do NOT add a new import. ONLY replace one line.

Output exactly in this format (two lines):
Original import: <the exact original line>
Hallucinated import: <the fake replacement line>
\end{promptbox}

\section{Judge Evaluation Prompt}
\label{app:judge_prompts}

The same system prompt is used for all four judges during curation:

\begin{promptbox}{Judge System Prompt}
You are a top-tier software engineer with deep expertise in software development and code review. Evaluate a candidate code completion that should connect the provided prefix and suffix inside the same file.

Score 1 ONLY when all of the following are true:
- The completion keeps the file syntactically valid.
- The completion is logically consistent with the surrounding context.
- The completion is relevant: it advances the task from the prefix toward the suffix.
- The completion does not hallucinate APIs, data, or behavior conflicting with the context.
Otherwise score 0.

Format your response exactly as:
Detailed Reasoning: <reasoning text>
Final Average Score for Completion: <0 or 1>
\end{promptbox}

\section{Execution Verification: Full Details}
\label{app:docker_full}

\paragraph{Verification protocol.}
For each sample, we construct two complete source files by inserting the golden or hallucinated completion between the prefix and suffix. The golden file must compile/parse and run without errors. The hallucinated file must produce the specific error type corresponding to its hallucination category: \texttt{AttributeError} (method), \texttt{TypeError} (parameter), \texttt{NameError} (undefined variable), or \texttt{ImportError} (import). Samples failing either condition are excluded.

\paragraph{Docker infrastructure.}
We maintain seven per-language base images, each equipped with the full toolchain and standard package manager:
Python (CPython 3.11 / pip), Java (OpenJDK 17 / Maven), TypeScript (Node.js 20 / npm), Go ($1.21$ / go modules), Rust (nightly / cargo), C++ (GCC 13 or Clang / apt), and C\# (.NET $8.0$ / NuGet). Each image includes a dependency-resolution module that analyzes the source file's imports and installs the required packages before verification.

\paragraph{Standalone-file precheck.}
\benchmark targets self-contained, file-level FIM evaluation, so verification is preceded by a static precheck that discards any source file whose dependency graph cannot be closed within the file itself: files that import sibling modules from the same repository, depend on project-internal build artefacts, or require harness code outside the file are rejected before any container is built. The precheck retains roughly $5\%$ of mined files, and this filter---rather than the verification stage that follows---is the dominant determinant of overall yield. The constraint is deliberate: it keeps every per-sample container lightweight and reproducible at the cost of excluding repository-level FIM scenarios, which are complementary and better served by repository-level benchmarks (see Appendix~\ref{app:discussion}).

\begin{figure}[h]
\centering
\includegraphics[width=0.85\linewidth]{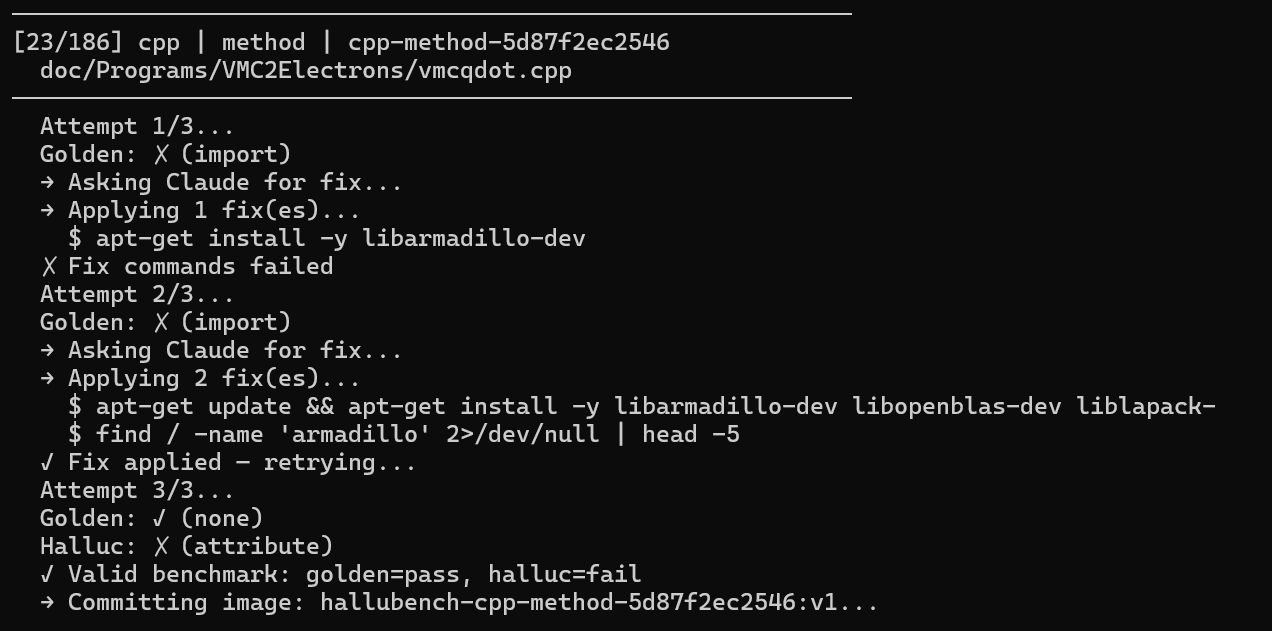}
\caption{\claude-assisted fix loop: a C++ sample requires three iterations to resolve missing dependencies before golden (pass) and hallucinated (expected fail) verifications both succeed.}
\label{fig:claude_debug}
\end{figure}

\paragraph{\claude-assisted fix loop.}
Real-world GitHub code frequently has external dependencies that are not immediately satisfiable in a clean container even after the precheck (system libraries, build flags, native toolchain configurations). Conditional on passing the precheck, roughly $0.4$ of the assembled (golden, hallucinated) tuples verify successfully on the first attempt. For the remainder, we run a \claude-assisted debugging loop: the error output is sent to \claude, which proposes a fix (e.g., installing missing system libraries, adding package dependencies, or adjusting build configurations); the proposed fix is applied to the container environment and verification is retried. The loop runs up to three iterations per sample and raises the verification rate among precheck-passing files to $\sim\!0.75$. Figure~\ref{fig:claude_debug} walks through one such case.

\paragraph{Container finalization.}
Successfully verified samples are packaged as self-contained Docker containers with all code and dependencies baked in, then pushed to Azure Container Registry. Each container supports three invocation modes: \texttt{verify golden} (confirm golden compilation), \texttt{verify hallucinated} (confirm error production), and \texttt{verify patch} (evaluate a model-generated completion via stdin). This interface enables drop-in integration with any FIM model evaluation pipeline.

\begin{dockerbox}
docker pull <registry>/<image_tag>

# Verify golden completion compiles
docker run <image> verify golden

# Verify hallucinated completion produces expected error
docker run <image> verify hallucinated

# Evaluate a model-generated completion (via stdin)
echo "<completion>" | docker run -i <image> verify patch

# Evaluate from a file (via volume mount)
docker run -v ./comp.txt:/input/patch.txt <image> verify patch

# Retrieve sample metadata
docker run <image> info       # JSON with all metadata
docker run <image> prompt     # FIM prefix context
docker run <image> golden     # correct completion text
docker run <image> hallucinated  # incorrect completion text
\end{dockerbox}

\texttt{verify patch} returns structured JSON: \texttt{\{"is\_valid": true|false, "error\_message": "..."\}}. Container images average $150$--$400$\,MB; containers run with \texttt{-{}-network=none} (except C\# which requires NuGet restore).

\section{Statistics: Full Details}
\label{app:stats_full}

Marginal language distribution and per-language complexity are reported in Tables~\ref{tab:lang_dist} and~\ref{tab:perlang_complexity}.

\begin{table}[h]
\caption{Language distribution.}
\label{tab:lang_dist}
\centering\small
\begin{tabular}{lrr}
\toprule
\textbf{Language} & \textbf{N} & \textbf{\%} \\
\midrule
TypeScript & 420 & 21.5 \\
Python & 374 & 19.2 \\
Go & 291 & 14.9 \\
Rust & 252 & 12.9 \\
C\# & 246 & 12.6 \\
Java & 243 & 12.5 \\
C++ & 125 & 6.4 \\
\midrule
\textbf{Total} & \textbf{1,951} & \\
\bottomrule
\end{tabular}
\end{table}

\begin{table}[h]
\caption{Per-language complexity metrics for \benchmark.}
\label{tab:perlang_complexity}
\centering\small
\begin{tabular}{lrrrrrrr}
\toprule
\textbf{Language} & \textbf{N} & \textbf{Pfx LOC} & \textbf{Cmp LOC} & \textbf{Tot LOC} & \textbf{Pfx Tok.} & \textbf{Cmp Tok.} & \textbf{CC} \\
\midrule
C++ & 125 & 42.5 & 1.0 & 83.5 & 319.2 & 7.5 & 11.3 \\
C\# & 246 & 110.2 & 2.1 & 255.3 & 1118.5 & 20.7 & 22.1 \\
Go & 291 & 97.6 & 1.5 & 210.4 & 945.7 & 12.9 & 24.9 \\
Java & 243 & 59.1 & 1.1 & 123.8 & 426.2 & 10.7 & 16.0 \\
Python & 374 & 42.9 & 1.7 & 152.1 & 344.8 & 11.8 & 13.8 \\
Rust & 252 & 65.0 & 1.6 & 174.5 & 540.3 & 13.1 & 19.2 \\
TypeScript & 420 & 30.8 & 3.7 & 77.0 & 203.4 & 25.8 & 9.6 \\
\midrule
\textbf{Overall} & \textbf{1,951} & \textbf{61.7} & \textbf{2.0} & \textbf{152.6} & \textbf{534.4} & \textbf{15.8} & \textbf{16.4} \\
\bottomrule
\end{tabular}
\end{table}

Per-language complexity varies considerably within \benchmark: C\# samples average $255.3$ LOC and $22.1$ CC (enterprise-style code), while TypeScript averages $77.0$ LOC and $9.6$ CC (compact web-development patterns). This variation ensures the benchmark tests hallucination resilience across diverse complexity profiles.

\section{Evaluation: Metrics, Per-Type, and Fine-Grained Results}
\label{app:metrics}\label{app:fine_grained}

\paragraph{Metric definitions.}
\textbf{pass@1} (execution-based): a completion \emph{passes} if the assembled file compiles and runs without errors when the prediction is inserted.
\textbf{Edit Similarity (ES)}: $1 - \text{Levenshtein}(\hat c, c_g)/\max(|\hat c|, |c_g|)$.
\textbf{CodeBLEU}~\citep{ren2020codebleu}: equally-weighted combination of n-gram, weighted n-gram, AST, and dataflow matching, using language-specific parsers.
\textbf{Hallucination Rate (HR)}: $\frac{1}{N}\sum_i \mathbb{1}[\text{sim}(\hat c_i, c_{h,i}) > \text{sim}(\hat c_i, c_{g,i})]$ with SequenceMatcher similarity.

\paragraph{pass@1 by hallucination type.}
\begin{table}[h]
\caption{pass@1 by hallucination type, Qwen2.5-Coder slate, computed on the $950$ unique FIM contexts. Import is consistently the most challenging across all model sizes.}
\label{tab:by_type}
\centering\small
\begin{tabular}{lrrrrrr}
\toprule
\textbf{Type} & \textbf{0.5B} & \textbf{1.5B} & \textbf{3B} & \textbf{7B} & \textbf{14B} & \textbf{32B} \\
\midrule
Import & .378 & .540 & .653 & .630 & .736 & .771 \\
Method & .474 & .695 & .766 & .747 & .831 & .851 \\
Parameter & .428 & .723 & .855 & .766 & .906 & .906 \\
Undef.\ Var. & .491 & .709 & .858 & .759 & .851 & .900 \\
\bottomrule
\end{tabular}
\end{table}

\begin{figure}[h]
\centering
\includegraphics[width=0.95\linewidth]{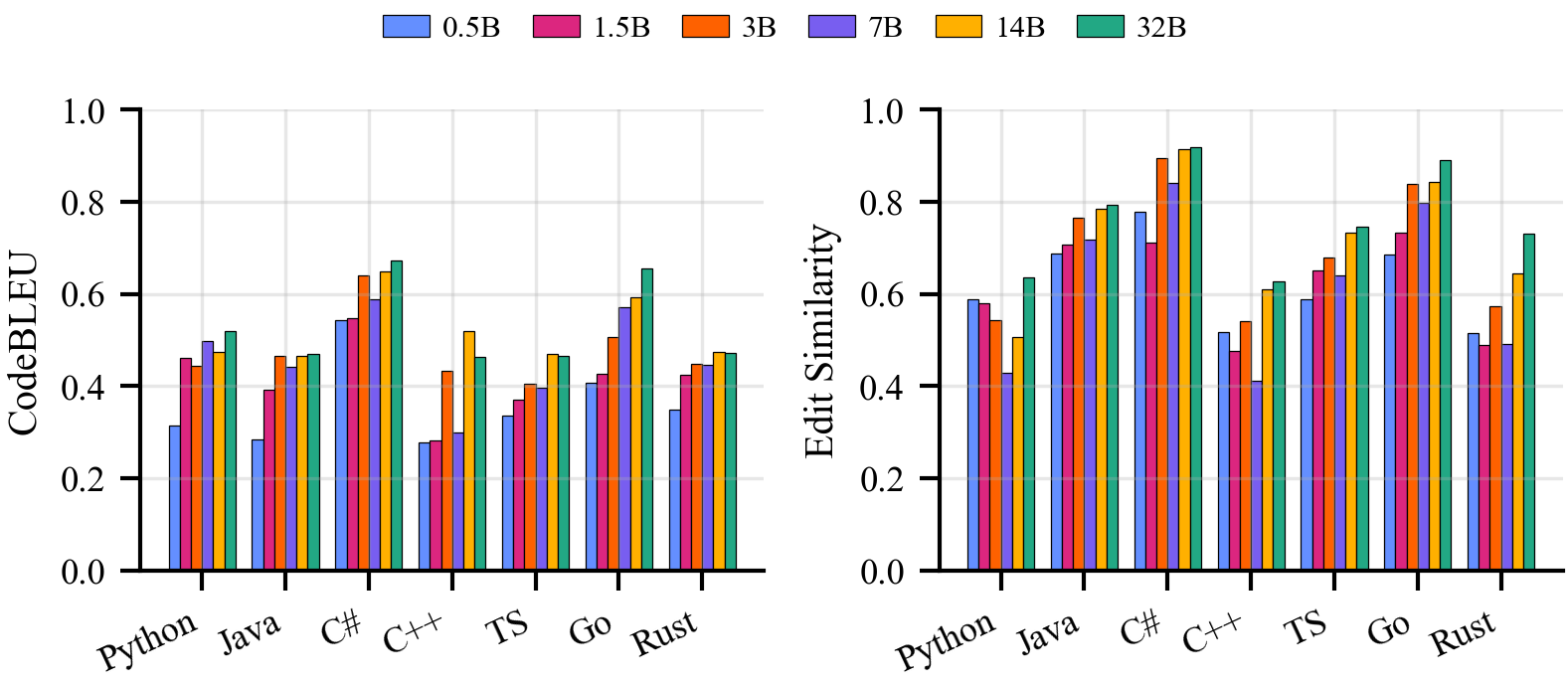}
\caption{CodeBLEU (left) and Edit Similarity (right) by language across Qwen2.5-Coder model sizes. Larger Qwen sizes generally improve both metrics; Qwen-0.5B's Edit Similarity is inflated by the line-budget cap and should be compared against pass@1 rather than read as an indicator of completion quality.}
\label{fig:quality}
\end{figure}

\paragraph{Code-quality metrics by language.}
Figure~\ref{fig:quality} reports CodeBLEU and Edit Similarity by language. Both correlate positively with pass@1 across the slate, but the correlation is not perfect: a model can produce a completion that is syntactically similar to the golden code yet functionally incorrect (or vice versa)---motivating our multi-metric approach. Note that Qwen-0.5B's Edit Similarity is unusually high relative to its pass@1 because the line-budget cap (\S\ref{sec:eval}) clips the most degenerate suffixes for that model only, leaving short well-formed prefixes that score generously on character-level similarity even when the underlying API knowledge is wrong.

\begin{figure}[h]
\centering
\includegraphics[width=0.95\linewidth]{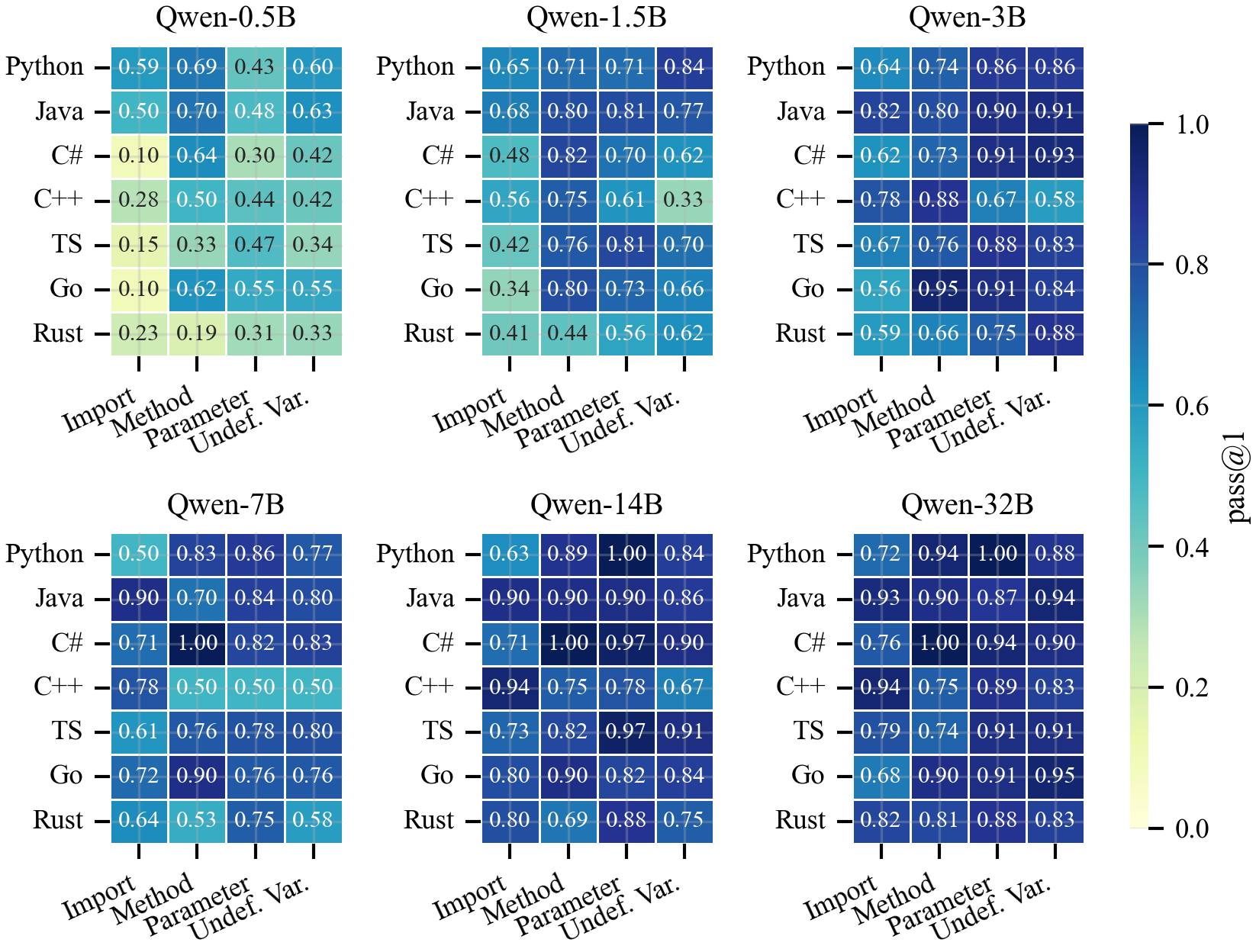}
\caption{pass@1 heatmaps per model: language (rows) $\times$ hallucination type (columns).}
\label{fig:heatmaps}
\end{figure}

\paragraph{Per-language and per-(language $\times$ type) heatmaps.}
Figure~\ref{fig:heatmaps} shows pass@1 heatmaps per model with language (rows) $\times$ hallucination type (columns), computed on the $950$ unique FIM contexts. Several patterns emerge: (i)~\textbf{Import across all languages} is consistently the weakest column at every scale, with C\# and Go imports proving especially challenging---Qwen-32B reaches only $0.76$ on C\#$\times$Import and $0.68$ on Go$\times$Import; (ii)~\textbf{C++ method/parameter cells} remain unstable across scales (the small $32$ unique-context C++ cell is sensitive to a handful of failures, e.g.\ $0.75$ on Method at both $14$B and $32$B); (iii)~\textbf{$0.5$B is uniformly poor} across all cells; (iv)~improvement with scale concentrates on specific (language, type) pairs rather than improving uniformly, and the $3$B$>$$7$B inversion (\S\ref{sec:eval}) is visible cell-by-cell---several cells regress between $3$B and $7$B (e.g.\ Python$\times$Param., C\#$\times$Param., Go$\times$Method) before recovering at $14$B/$32$B.

\section{Discussion and Limitations}
\label{app:discussion}

\subsection{Benchmark difficulty and validity.}
Four pieces of evidence support the claim that \benchmark is genuinely challenging rather than a pattern-matching test. (1)~The adversarial mining loop is empirically effective: cross-judge both-correct accuracy on the unverified Iter-0 pool decreases monotonically across iterations (Table~\ref{tab:judge_iter}), confirming that progressively harder samples are surfaced. (2)~On the final execution-verified release, even the strongest detection judge (Claude-4.5-Opus) achieves $92.1\%$ both-correct accuracy (Table~\ref{tab:detection}), with import hallucinations remaining the hardest category at $83.4\%$; detection is therefore unsolved on \benchmark even for frontier judges. (3)~Docker-based execution verification guarantees that every retained hallucination produces the categorical runtime error class, removing the LLM-label noise that inflated the Iter-0 numbers. (4)~The strongest of the $11$ generators we evaluate (Qwen2.5-Coder-32B-Instruct) reaches $84.5\%$ pass@1 (on the $950$ unique FIM contexts) and no cross-family model exceeds $0.77$ Edit Similarity, so neither the judge frontier nor the FIM-generator frontier solves the benchmark. The two-stage human-review pass (Appendix~\ref{app:human_review}; $1{,}957$ Docker-verified candidates $\to$ $1{,}951$ released, with $207$ edits re-verified through Docker) further removes residual single-generator artefacts that could otherwise be confused with intrinsic difficulty.

\subsection{Scaling behavior and capability emergence.}
The transition from $0.5$B ($43.6\%$) to $1.5$B ($64.7\%$) suggests a \emph{capability threshold}: below $1.5$B parameters, models lack the minimum semantic understanding needed for reliable FIM completion in hallucination-prone contexts---and even after the line-budget cap, the $0.5$B model only matches the gold completion byte-for-byte on $22.8\%$ of samples and reaches an Edit Similarity of $0.625$, well below every other Qwen size. The $3$B$>$$7$B inversion (\!$76.7\%$ vs.\ $71.1\%$, with $7$B showing \emph{lower} Edit Similarity ($0.625$ vs.\ $0.693$) and Exact Match ($0.402$ vs.\ $0.440$) than $3$B) is reproducible across all four metrics in our single-seed run; we report it as an open anomaly and decline to attribute it to a specific cause without multi-seed verification, layer-wise probing, or training-recipe details that are not publicly documented (\S\ref{sec:eval}). The modest gap between $14$B ($81.5\%$) and $32$B ($84.5\%$) suggests that reducing hallucination susceptibility beyond a certain scale may require targeted interventions---training-data curation, post-training alignment on hallucination-specific signals, or specialized fine-tuning---rather than simply increasing model size.

\subsection{Hallucination type difficulty hierarchy.}
Across the Qwen scaling slate (Table~\ref{tab:by_type}) and the cross-family slate, \emph{import is consistently the hardest category}: at $32$B, pass@1 on Import is $0.771$ versus $0.851$--$0.906$ on the other three categories, and the same ordering holds at every smaller scale. The remaining three categories (Method, Parameter, Undef.\ Var.) cluster within a few percentage points of each other at $32$B and we therefore describe them as statistically indistinguishable on the present sample sizes. The Import gap is plausibly explained by the semantic demand of each category: Import requires memorized knowledge of the actual package ecosystem (knowing that \texttt{sklearn.neural} does not exist demands traversing the ecosystem hierarchy), whereas Method, Parameter, and Undef.\ Var.\ admit at least partial inference from local context---surrounding identifiers, type annotations, or scope. Per-cell confidence intervals would sharpen the within-tier comparisons.

\subsection{Dual use: generation and detection.}
\benchmark serves two purposes: (i)~\emph{generation evaluation}---running model inference on \benchmark and measuring pass@1 quantifies how often a FIM model produces hallucinated completions; (ii)~\emph{detection evaluation}---the judge framework from our curation pipeline can be repurposed to evaluate how well models detect hallucinations in suggested code, increasingly important for code-review assistants and agentic coding systems. The paired golden/hallucinated completions, together with difficulty scores, also provide a potential training signal for supervised or preference-based fine-tuning.

\subsection{Limitations.}
\begin{itemize}[leftmargin=*]
    \item \textbf{Sample size.} The released artifact contains $1{,}951$ samples ($\sim\!280$ per language, $\sim\!70$ per (language, type) cell on average); cells below $\sim\!50$ samples (notably Python$\times$Parameter at $22$ and C++ at $125$ overall) carry visibly wider sampling intervals and per-cell numbers should be read as suggestive rather than confirmatory. An extended release in which every (language, hallucination type) cell contains at least $100$ verified samples is currently in curation.

    \item \textbf{Standalone-file constraint.} To keep each per-sample Docker container lightweight, we admit only source files whose static dependency graph is closable within the file itself (\S\ref{sec:verification}). Samples that genuinely require multi-file or repository-level context are therefore under-represented; \benchmark complements rather than replaces repository-level FIM benchmarks such as CrossCodeEval~\citep{dingCrossCodeEvalDiverseMultilingual2023}.
   
    \item \textbf{Hallucination taxonomy.} The 4 categories cover the most common knowledge-level FIM hallucination modes from~\citet{gao2025hallucination_survey} but exclude logic errors, type-mismatch errors that compile silently, deprecated-API usage, behavior-correct-but-API-wrong completions, and other semantic-correctness failures that do not raise a categorical runtime error. \benchmark should accordingly be read as a \emph{lower bound} on hallucination prevalence (\S\ref{sec:taxonomy}).
    \item \textbf{Completion length.} Completions average $2.0$ LOC, reflecting the typical granularity of localized FIM hallucinations targeted by our taxonomy. Multi-statement and block-level hallucinations would test different aspects of hallucination resilience and are not in scope for this release.
    
    \item \textbf{Temporal validity.} Package ecosystems evolve; some hallucinated imports may become valid (or golden ones may be deprecated) over time. The released artifact is pinned to the October 2025 mining snapshot; we plan periodic refreshes alongside the extended release.
    
    \item \textbf{Intended use.} \benchmark is released as a \emph{test-only} artifact with no pre-defined train/validation split. Fine-tuning on \benchmark samples is explicitly discouraged. Researchers needing a training source for hallucination mitigation should regenerate samples through the released pipeline against a different seed corpus.
    
    \item \textbf{Misuse.} The paired golden/hallucinated structure could in principle be used to fine-tune models that produce \emph{more} convincing hallucinations. We consider this risk modest relative to the benefit of measurable hallucination evaluation, and we mitigate it operationally by releasing only the curated benchmark (not the intermediate generation/judging pipeline outputs) and by recommending that any defensive fine-tuning on \benchmark be reported transparently.
\end{itemize}

\section{Qualitative Examples}
\label{app:examples}

We present one representative sample per hallucination type. Each example shows the FIM context, the golden (correct) completion, and the hallucinated (incorrect) completion with its resulting error.

\subsection*{Import Hallucination.}

\begin{codebox}{Context (prefix tail)}
import os, sys
from pathlib import Path
\end{codebox}
\begin{goldenbox}
from PIL import Image, ImageFilter
\end{goldenbox}
\begin{hallucbox}
from Pillow import Image, ImageFilter
\end{hallucbox}
{\small\color{red!70!black}\texttt{ModuleNotFoundError: No module named `Pillow'}}

\subsection*{Method Hallucination.}

\begin{codebox}{Context (prefix tail)}
@app.route('/fact/<fact>',
           defaults={'env': ..., 'value': None})
@app.
\end{codebox}
\begin{goldenbox}
route('/<env>/fact/<fact>', defaults={'value': None})
\end{goldenbox}
\begin{hallucbox}
endpoint('/<env>/fact/<fact>', defaults={'value': None})
\end{hallucbox}
{\small\color{red!70!black}\texttt{endpoint() got an unexpected keyword argument `defaults'}}

\subsection*{Parameter Hallucination.}
\begin{codebox}{Context (prefix tail)}
sa.column('name', sa.String),
sa.column('abbr', sa.String),
sa.
\end{codebox}
\begin{goldenbox}
column('service', sa.String)
\end{goldenbox}
\begin{hallucbox}
column('service', sa.String, nullable=False)
\end{hallucbox}
{\small\color{red!70!black}\texttt{column() got an unexpected keyword argument `nullable'}}

\subsection*{Undefined Variable Hallucination.}
\begin{codebox}{Context (prefix tail)}
class JudgeDecisionSchema(BaseModel):
    """Schema for judge decisions"""
    approved: bool =
\end{codebox}
\begin{goldenbox}
Field(..., description="Whether the card is approved")
\end{goldenbox}
\begin{hallucbox}
Field(..., description=approval_default)
\end{hallucbox}
{\small\color{red!70!black}\texttt{NameError: name `approval\_default' is not defined}}
\section{Source Data Provenance}

The benchmark draws from a corpus of ${\sim}1.1$ million code completion examples mined in October 2025 from ${\sim}11{,}975$ public GitHub repositories across 9 languages and 464 third-party APIs. For each language, ${\sim}25$ target packages were selected by a consensus of three signals: (i) adoption frequency in anonymized code completion telemetry, (ii) LLM-based ecosystem ranking, and (iii) curated priority lists from language-specific engineering teams. Each target package was pinned to a minimum version released within a 4-month window of the mining date; only repositories declaring a dependency at or above that version were retained. C++ libraries were selected via expert curation and matched by include-path patterns, as C++ lacks a standardized package manager. Source files were filtered to the 2nd--98th percentiles for file size and line count. API call sites were extracted using tree-sitter AST parsing, classified by GPT-4o-mini, and split into fill-in-the-middle (prefix, completion, suffix) triplets at the call site's byte offsets.

From this corpus, 1{,}951 examples were sampled across 7 languages and 319 repositories. All examples are drawn exclusively from public GitHub repositories. We traced each example to its source repository and retrieved metadata. Across the 319 repositories, the median GitHub star count is 81, with a range of 3 to 76{,}366 (mean: 1{,}890); 56\% have $\geq$50 stars and 22\% have $\geq$1{,}000 stars. Each released sample includes the source repository URL and detected license metadata for provenance purposes.

For languages with standardized package managers, the source corpus retains only repositories that declared dependencies on package versions released within 4 months of the October 2025 mining date. Table~\ref{tab:freshness} reports the earliest target package release date per language.

\begin{table}[h]
\centering
\caption{Earliest target package release date per language at the time of mining (October 2025).}
\label{tab:freshness}
\begin{tabular}{lll}
\toprule
\textbf{Language} & \textbf{Earliest Package Date} & \textbf{Age at Mining} \\
\midrule
C\#        & April 2025 & ${\sim}6$ months \\
Java       & May 2025   & ${\sim}5$ months \\
PHP        & June 2025  & ${\sim}4$ months \\
Python     & June 2025  & ${\sim}4$ months \\
Rust       & June 2025  & ${\sim}4$ months \\
Go         & July 2025  & ${\sim}3$ months \\
TypeScript & July 2025  & ${\sim}3$ months \\
JavaScript & July 2025  & ${\sim}3$ months \\
C++        & N/A (curated list) & --- \\
\bottomrule
\end{tabular}
\end{table}